\documentclass[conference]{IEEEtran}
\IEEEoverridecommandlockouts
\usepackage{graphicx} 
\usepackage[dvipsnames]{xcolor}
\usepackage{amsmath}
\usepackage{amsfonts}
\usepackage{amssymb}
\usepackage{algorithm}
\usepackage{algorithmic}

\usepackage{subcaption}

\usepackage{float} 
\usepackage{multicol}
\usepackage{fvextra} 

\usepackage{array}
\newcolumntype{P}[1]{>{\centering\arraybackslash}p{#1}}

\title{Improving Medical Diagnostics with Vision-Language Models: Convex Hull-Based Uncertainty Analysis}

\author{\IEEEauthorblockN{Ferhat Ozgur Catak\(^1\) , Murat Kuzlu\(^2\), Taylor Patrick\(^2\)}
\IEEEauthorblockA{
\textit{\(^1\)Department of Electrical Engineering and Computer Science, University of Stavanger}, Rogaland,
Norway \\ 
\textit{\(^2\)Batten College of Engineering and Technology, Old Dominion University}, Norfolk, VA, USA\\
f.ozgur.catak@uis.no, mkuzlu@odu.edu, tpatr004@odu.edu}
\\ Corresponding author: Ferhat Ozgur Catak f.ozgur.catak@uis.no}

\begin{document}

\maketitle
\begin{abstract}

In recent years, vision-language models (VLMs) have been applied to various fields, including healthcare, education, finance, and manufacturing, with remarkable performance. However, concerns remain regarding VLMs' consistency and uncertainty, particularly in critical applications such as healthcare, which demand a high level of trust and reliability. This paper proposes a novel approach to evaluate uncertainty in VLMs' responses using a convex hull approach on a healthcare application for Visual Question Answering (VQA). LLM-CXR model is selected as the medical VLM utilized to generate responses for a given prompt at different temperature settings, i.e., 0.001, 0.25, 0.50, 0.75, and 1.00. According to the results, the LLM-CXR VLM shows a high uncertainty at higher temperature settings. Experimental outcomes emphasize the importance of uncertainty in VLMs' responses, especially in healthcare applications. 

\end{abstract}


\section{Introduction}

In recent years, the use of artificial intelligence (AI) has led to the development of large language models (LLMs) and vision language models (VLMs), with remarkable performance. These technologies have been extended to multimodal LLMs (such as GPT-4V \cite{achiam2023gpt}, LLaVA \cite{liu2024improved}, CogVLM \cite{wang2023cogvlm}, CLIP \cite{radford2021learning}), which enhance problem-solving capabilities by evaluating text, audio/speech, images, and videos. Multimodal LLMs also have the potential to be applied to medical and clinical scenarios to improve classification, question answering, informed decision making, efficiency, educational methods, patient care, and minimize medical mistakes \cite{cirone2024assessing}. Models such as GPT-4 with vision (GPT-4V)\footnote{https://openai.com/index/gpt-4v-system-card/} are large-scale multimodal models, which can accept image and text inputs and produce text outputs \cite{achiam2023gpt}. GPT-4 is a transformer-based model pre-trained to predict the next token in a document exhibiting human-level performance on various professional and academic benchmarks \cite{achiam2023gpt}. GPT-4 has also shown promise in medical and clinical tasks. Guerra et al. found that GPT-4 outperforms ChatGPT, medical students, and neurosurgery residents on neurosurgery written board-like questions \cite{guerra2023gpt}. Zhou et al. examined OpenAI's generative pre-trained transformer with vision potential (GPT-4) for automated image text pair generation, noting that it has shown promise in understanding natural images, but had limited effectiveness in interpreting real-world chest radiographs \cite{zhou2024evaluating}.
%
%

Llama is a collection of pre-trained and finetuned (chat and dialogue cases) foundational LLMs ranging in scale from 7 to 70 plus billion parameters. Some foundational LLMs such as GPT-4 and Llama (versions 1 through 3) \cite{touvron2023llama, touvron2023llama2} have been adapted to function as VLMs, with existing vision models, to facilitate multimodal predictions and generations.

VLMs are models that utilize both image and text information to perform complex reasoning tasks and human-level language comprehension for enhanced decision-making support compared to unimodal models. VLMs are often configured with fused individual unimodal vision and language AI models to perform multimodal classifications, predictions, and/or generations given an input of image and/or text \cite{ferraro2015survey}. By mimicking the multimodal nature of clinical expert decision-making, VLMs can significantly enhance medical diagnosis and decision-making through improved predictive performance utilizing multimodal health information (signs, symptoms, imaging, written reports, physiological and laboratory measurements) \cite{capobianco2023assessment}. 

Ideally, VLMs aim to achieve expert human-level functioning, as medical tasks are challenging without the use of AI/ML or computer assistance due to image-text complexity, variability, noise, and resolution. LLM extensions to VLMs have been explored for medical and clinical tasks and applications. For example, Wang et al. developed DRG-LLaMA \cite{wang2024drg}, which tuned LLama to predict the diagnosis-related group for hospitalized patients and found that performance was correlated with increased model parameters and input context lengths.
Additionally, Sandmann et al. performed a systematic analysis of ChatGPT, Google search, and Llama 2 for clinical decision support tasks \cite{sandmann2024systematic}. Progress in LLMs has made the generation of realistic image caption tasks viable and expansive. However, these models often struggle to make accurate, accurate, relevant, and consistent statements, which in turn negatively affects their trustworthiness and reliability \cite{zohar2024lovm}. This is especially true of non-fine-tuned models. The diagnostic accuracy and interpretability of the medical image and report models are key to accurate medical analysis, diagnosis, and subsequent treatment and care. More importantly, the uncertainty of unsafe suggestions by any model, including VLMs, are important to quantify before use in medical or clinical settings.


\section{Preliminaries}
\subsection{Uncertainty and Consistency}

Uncertainty quantification (UQ) has received more attention in the context of generative AI (GAI), particularly critical applications using LLMs in healthcare \cite{kostumov2024uncertainty, groot2024confidence, tharayil2024framework, catak2024trustworthy}. Uncertainty can originate from various factors such as the model's architecture, the parameters that define the model, the dataset \cite{nemani2023uncertainty, jalaian2019uncertain}, and the overall performance of LLM/VLM. Training data can also contribute significantly to the uncertainty as a result of the complexity and diversity of the dataset.

VLMs often struggle to provide accurate or true conclusions and representations on tasks. This low performance is due to the improper analysis and comprehension of information from multiple modalities. 
VLMs that perform Vision Question Answering (VQA) tasks have been shown to lack robustness and are severely prone to overfitting on dataset-specific correlations rather than learning to answer questions \cite{khan2024consistency}. VQA models often use simple rules, based on co-occurences of objects with noun phrases and linguistic priors, to answer questions (e.g., the fox is red) rather than referring to the image for context (e.g., the fox is white) \cite{khan2024consistency}. VLMs may also override visual information and substitute or prioritize prior learned (visual and language) information due to co-association.
This phenomenon is referred to as poor visual grounding, meaning that VLM inferences and information from one modality are prioritized over the other modality(s) and as a result the performance of the model often suffers \cite{selvaraju2019taking}.
Therefore, cross-modal alignment, multimodal attention, and prioritization are important concepts when evaluating multimodal consistency and hallucination.
Inconsistency represents the uncertainty and confusion of the model towards a given task, to be a contributing factor to various types of hallucination in language and vision-based models \cite{zhang2024evaluating, zohar2024lovm}. 

Khan et al. hypothesized that VQA models answer simpler questions more consistently, with VQA task inconsistency on linguistic variations being often indicative of a more superficial understanding of the question content \cite{khan2024consistency}. As a result, the answer provided is more likely to be wrong or factually incorrect.
Since consistency and confidence have been shown to not be equivalently related with respect to questions and answers, the predictive uncertainty of the model can be quantified using consistency instead of accuracy-based predictions \cite{khan2024consistency}. Uncertainty can be quantified by metrics such as the Attribution Based Confidence (ABC) metric, based on the feature attribution guidance, which uses specifically integrated gradients to perturb samples in a feature space, and then evaluates consistency over the perturbed samples \cite{jha2019attribution}.
In a black box model, where features are inaccessible, there is often no direct way to explore the input neighborhood in a feature space \cite{khan2024consistency}.
Often only the raw confidence scores for answer candidates are available with black box models, as a result the confidence of the most likely answer may be utilized as the uncertainty \cite{khan2024consistency}.
Alternatively, when features are inaccessible, rephrasing can be applied where an alternate surface form of the input is mapped closely to the original input in feature space \cite{khan2024consistency}.
Khan et al. have found that consistency in rephrasing is an effective step in evaluating black-box VQA models for predictive uncertainty, especially when the answers of queries are unknown \cite{khan2024consistency}.  

It is important to understand how similar two QA responses are to each other, with regard to content and understanding, to determine how consistent or uncertain a model is \cite{zohar2024lovm}.   
Language models often employ decoding strategies to improve language generation quality, such as re-ranking, temperature sampling, top-k sampling, top-p sampling, nucleus sampling, typical decoding, and minimum Bayes risk decoding \cite{wang2022self}. 
%
%
Self-consistency is applicable to improving the performance of a wide range of reasoning tasks without any additional supervision, training, data collection nor finetuning. 
Wang et al. determined, for a given model, the optimal answer by marginalizing out sampled reasoning paths to find the most consistent answer in the final answer set \cite{wang2022self}. Self-consistency avoids the repetitiveness and local optimality of greedy decoding algorithm methods, while mitigating the stochasticity of a single sampled generation \cite{wang2022self}. 
Consistency and self-consistency can be extended to open-ended text generation tasks. This is possible if a good consistency metric can be defined between multiple generations of text, i.e., whether the answers agree or contradict \cite{zhang2024evaluating}. According to Zhang et al., multiple types of consistency exist that can affect the model, such as inner and outer consistency \cite{zhang2024evaluating}. 
Inner inconsistency refers to a model responding ‘yes’ to even contradictory questions. As a result, it is unclear whether the model accurately comprehends the truth of the ground or exhibits confusion, thus contributing to hallucination \cite{zhang2024evaluating}.  
Outer inconsistency refers to a model responding ‘no’ to its own answer, and as a result it conflicts with itself, which is inconsistent.  
This outer inconsistency further reveals the uncertainty of the model about the query and may contribute to hallucination \cite{zhang2024evaluating}. 
Inner and outer consistency can be utilized to evaluate the performance of various language tasks such as binary classification questions (yes/no/counting) or comparison questions, but may not fully capture the model's ability to answer open-ended questions \cite{zhang2024evaluating}. 
Therefore, we can achieve a more comprehensive understanding of model uncertainty, reliability, and hallucination by analyzing multiple types of consistency for model outputs \cite{zhang2024evaluating}. 
%

%

Prior studies leveraged synonyms to evaluate LLMs and can be extended to text generation tasks for VLMs where prompts are used to generate a list of semantically similar synonyms for every object class \cite{zohar2024lovm}. For example, a LLM/VLM pre-trained on instances of the chair class referenced with synonyms such as (chair, seat, couch, etc.) would be expected to have embeddings closely located in a shared embedding space. Recent studies show that synonym consistency can be utilized in language tasks to correlate the degree of familiarity or awareness of the model with a particular concept. A high synonym score between the class and its corresponding synonyms indicates the model is aware of semantic meaning of the class and more likely to have higher consistency and lower uncertainty on similar tasks.
%

%

\subsection{Temperature Setting and Sensitivities}

Vision-language models (VLMs) use visual and textual datasets to generate content combining image and language modalities. Temperature settings and sensitivities during training and inference can significantly impact the performance of VLMs. The temperature setting is applied in the softmax function to regulate the sensitivity of the resulting probability distribution. Lower temperatures make the distribution more confident (peaky), while higher temperatures make it more uniform.
Techniques that alter diversity in language models for text generation tasks such as question answering, image captioning, open-ended answer dialogue, and machine translation must control the relative trade-off between quality and diversity \cite{zhang2021trading}.  
Decoding methods such as nucleus sampling, top-k and top-p sampling, and temperature sampling allow for control of model output diversity and quality \cite{zhang2021trading}. These methods can be quickly implemented on top of pre-trained language-based models. 
Temperature sampling "divides the logits of each token by the temperature hyperparameter before normalizing and converting the logits into sampling probabilities" to re-estimate the softmax distribution \cite{zhang2021trading, zhu2024hot}. 
This is often used in natural language generation to reshape the probability distribution by introducing a temperature coefficient T to control the level of sampling randomness for model uncertainty, robustness, and reliability tests \cite{zhu2024hot}.

\subsection{Convex Hull}
The convex hull of a set of points is a fundamental mathematical structure utilized in statistics and computational geometry \cite{avis1995good}. It is an important statistical problem with many applications in location-based services, computer vision (image processing, pattern recognition), robotic sensor databases, statistical analysis, and data mining \cite{yan2014probabilistic, agarwal2017convex, sirakov2006new}. 
The convex hull problem is meant to handle data uncertainty of individual points over a given area, with numerous existing algorithms that attempt to compute the probability of a query point lying inside the convex hull of the input. Considerations for convex hull solving algorithms include computational efficiency, time-space trade-offs, and effectiveness \cite{agarwal2017convex}.  
Statistical information can be utilized to find the best representation of the probability distribution of the query data point, specifically data in which the location and potentially the relative location is uncertain (and potentially changing) inside the convex hull.

%
The convex hull problem is often investigated under two models of uncertainty, unipoint and multipoint: the unipoint or tuple model, where each point has a fixed position but only exists with some probability (0 to 1), and the multipoint model with each point having multiple possible locations or not appearing at all \cite{suri2013most}. 
Some algorithms for determining parameters regarding the convex hull include variations of the gift-wrapping algorithm, divide and conquer algorithm, and incremental algorithm \cite{phan2015direct, leung1997neural}. 

In this study, a convex hull-based approach is used to evaluate the uncertainty of VLMs' responses for a selected
healthcare application using the VQA task.


\section{Experimental Setup Components}
This section provides a brief description of experimental setup components, including the VLM model (LLM-CXR), the temperature settings, and the chest X-ray dataset.

\subsection{LLM-CXR}



In this study, the LLM-CXR model \cite{lee2023llm} is utilized as the VLM, which is an Instruction Finetuned LLM for CXR Image Understanding and Generation. This multimodal model was developed for clinical and medical applications, specifically chest X-rays (CXRs). The LLM-CXR model can perform several VLM tasks including image captioning, visual question answering (VQA), natural language comprehension, and image generation. It was developed based on an approach introduced in previous work \cite{esser2021taming}, which features a transformer and the architectural component VQGAN combination for the generation of bidirectional images and texts.  
The group developed instruction fine-tuning methods for pre-trained LLMs to be modified to operate as a multimodal vision language model. The modification process produces a final VLM model capable of input and output in both text and image format but involves no modification of the original LLM model structure or objectives.  

Specifically, the LLM-CXR utilizes an image adapter module to tokenize image inputs. These image tokens are then fed into an LLM along with other word tokens. The LLM used involves a fine-tuned dolly-v2-3b model \cite{conover2023free}, which is fine-tuned for the instruction-following task based on the GPT-NeoX architecture, as a base model \cite{lee2023llm}. The output of the LLM, the text tokens, and features, are fed to another adapter combined with an image generative model capable of generating multiple modalities for multiple tasks. This model was trained on the MIMIC CXR JPG dataset \cite{johnson2019mimic}. 

For the case of this presented method, images are tokenized by VQGAN. VQGAN is frozen during LLM training for clinical information-preserving CXR tokenization. Then the token embedding space is expanded for the LLM for further training and fine-tuning. The next data augmentation was performed with a synthetic VQA to evaluate the pertinence of language comprehension and enhance vision language alignment using Llama 2 to generate CXR questions and answers for training the LLM-CXR. Finally, image text bidirectional instruction fine-tuning was applied to optimize the LLM to address the following tasks and experiments: 


\begin{itemize} 
\item \textbf{NL-IF task:}  Natural Language Instruction-Following (NL-IF) involves the use of the NL-IF dataset when finetuning the LLM-CXR to perform instruction following tasks. 
\item \textbf{CXR to report generation task:} Generate CXR reports given CXR images using LLM-CXR, and the following performance evaluation techniques for image understanding: CheXpert labeler model,  ROUGE L, METEOR, CIDEr. The similarity between reports and ground truth reports was evaluated using AUROC/F1 and Jaccard similarity. 
\item \textbf{CXR VQA task:} Ask questions about the presence, location, and severity of lesion or findings for each CXR image and notes using the MIMIC CXR dataset. 
\item \textbf{Report to CXR generation task:} LLM-CXR generates CXR images matching chest X-rays described in a given text report. The ground truth in this case is original CXR images from MIMIC CXR JPGs and was compared to generated images calculating AUROC/F1.  
\end{itemize} 

The overall results for each of the LLM-CXR tasks demonstrate comparable or better performance to similar models at the time of publication \cite{lee2023llm}.  
One major issue with the LLM-CXR model is that when the same inquiry (question) and image is repeatedly asked of the model, often inconsistent and potentially incorrect answers are provided. This paper explores a method to evaluate and potentially improve issues where a model does not provide equally valuable and similar answers when given the same input for the image and the text-based question. 

\subsection{Temperature Settings}

The transformer base class pre-trained configurations implement methods for loading/saving a pre-existing configuration from local or online library/repository. Each derived config class implements model specific attributes such as parameters linked to tokenizing, fine tuning tasks, and sequence generation. 

The temperature (T) is an optional positive value that typically defaults to 1.0. The temperature is used to model the next token probabilities that will be used by default in the generate method of the model. Temperature is one of the crucial parameters in both LLMs and VLMs, which affects creativity and accuracy, i.e.,, low temperature (0 or near 0) offers more precise and repetitive outputs, while high temperature ($\geq$1) offers more diverse and random outputs.

In this research, the temperature value is defined in the LLM-CXR model initiation function code in the ranges (0.001, 0.25, 0.5, 0.75, 1.00) applied for 30 trials per image in the chest radiograph dataset \cite{cohen2020covid}.

\subsection{Chest X-ray Dataset}
The chest X-ray dataset features a public open dataset of chest X-ray and computed tomography (CT) images of patients that are positive or suspected of COVID 19 or pneumonia (either viral or bacterial such as MERS, SARS, or ARDS) \cite{cohen2020covid}. Data were collected from public sources, hospitals and physicians and have been published in the corresponding GitHub repository\footnote{https://github.com/ieee8023/covid-chestxray-dataset/tree/master/images}. 

The final diagnosis for a given image can be found in the metadata CSV file, labeled under findings, which indicates the diagnoses type of lung disease or pneumonia that was given by medical workers and professionals. Other relevant information featured in the metadata CSV file includes patient ID, patient sex, age, vital signs, clinical notes, etc.

\section{Experimental Setup and Results}
This section describes the overall experimental setup and discusses the results for different temperature settings in terms of uncertainty.

Our codes are released on GitHub for scientific use.\footnote{https://github.com/ocatak/VLM\_Uncertainty}

\subsection{Experimental Setup and Uncertainty Evalaution}

\begin{figure*}[h]
    \centering
    \includegraphics[width=1.0\linewidth]{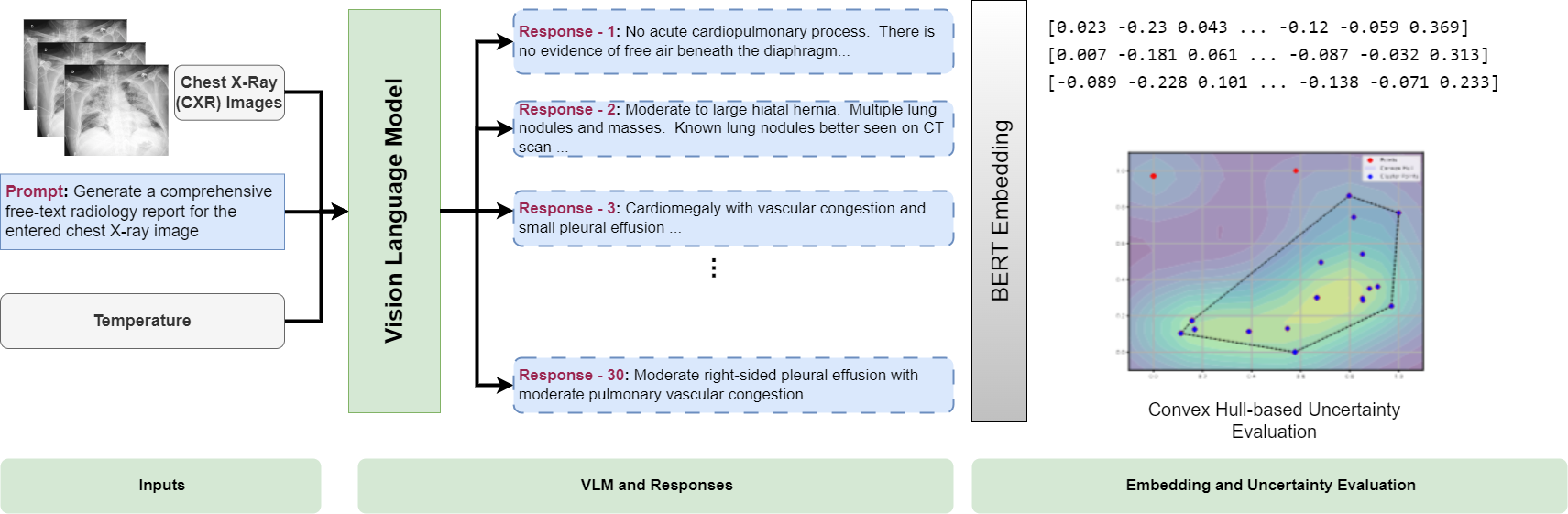}
    \caption{The overall experimental setup for calculating uncertainty in VLM responses.}
    \label{fig:system_overview}
\end{figure*}

The overview of the experimental setup for calculating the uncertainty in the VLM responses is shown in Figure \ref{fig:system_overview}, derived from the study \cite{ozgur2024uncertainty}. The figure illustrates the overall experimental setup, from the inputs of chest X-ray images to uncertainty evaluation of the responses based on a convex hull-based approach. The setup starts with three inputs, that is, multiple chest X-ray images, a given prompt ("Generate a comprehensive radiology report for the entered chest X-ray image.") to generate a radiology report, and a temperature setting. These inputs are processed by the selected VLM, i.e., LLM-CXR, to understand the visual content of the X-ray images and generate radiology reports. In this setup, 30 different responses were generated for each chest X-ray. The diversity in responses is controlled by a temperature input set to different values (0.001, 0.25, 0.50, 0.75, and 1.00), influencing the diversity of the reports generated. These responses are then encoded into high-dimensional embeddings using a BERT model. Then, the embeddings, initially in a high-dimensional space, are projected onto a 2D space using Principal Component Analysis (PCA) for easier visualization and clustering using the Density-Based Spatial Clustering of Applications with Noise (DBSCAN) algorithm, which identifies groups of similar responses. For each cluster, a convex hull is finally computed, representing the smallest convex boundary, i.e., the area of each convex hull, that encloses all points in the cluster. The total area is a measure of the uncertainty of the model's responses to the given prompt. The example shown includes a plot of the 2D embeddings with the convex hull of the densest cluster highlighted, illustrating the spatial distribution and clustering of the responses.

For the given prompt (\( p \in \mathcal{P} \)), the responses are generated, \( \mathcal{R}(p) = \{r_1, r_2, \ldots, r_n\} \), by the selected VLM, LLM-CXR, where $n=30$ is the number of responses.

The embeddings for each response \( r \in \mathcal{R}(p) \) are calculated using a pre-trained BERT model, and given as follows:
\[
\mathbf{E}(r) = \text{BERT}(r)
\]
where \( \mathbf{E}(r) \in \mathbb{R}^d \) represents the embedding vector in a \( d \) dimensional space, encapsulating the semantic content of the response within a high-dimensionalfeature space.

\(\mathbf{E}(\mathcal{R}(p)) = \{\mathbf{E}(r_1), \mathbf{E}(r_2), \ldots, \mathbf{E}(r_i)\} \) is the embedding set projected $d=2$ using Principal Component Analysis (PCA) to reduce the dimensionality of the vector for effective visualization and clustering, and given as follows:
\[
\mathbf{E}_\text{PCA}(\mathcal{R}(p)) = \text{PCA}(\mathbf{E}(\mathcal{R}(p)), 2)
\]

where \( \mathbf{E}_\text{PCA}(r) \in \mathbb{R}^2 \), the transformation is achieved by projecting the original embeddings onto a 2D subspace.

The DBSCAN algorithm is utilized to cluster the 2-D embeddings in order to detect distinct groups within the response space:
\[\mathbf{L} = \text{DBSCAN}(\mathbf{E}_\text{PCA}(\mathcal{R}(p)), \epsilon = 0.25 , \text{min\_samp}=3)\]\\
where \( \mathbf{L} \) represents the set of cluster labels assigned to each embedding point, with \(\epsilon \) controlling the maximum distance between points in the same cluster and \( \textit{min\_samp}\) specifying the minimum number of points required to form a cluster.

For each cluster \( c \in \mathbf{L} \) without noise points (\( c = -1 \)), the convex hull is calculated along with the corresponding area that encapsulates the geometric boundary of the cluster:
\[
\text{ConvexHull}(\mathbf{E}_\text{PCA}(c)), \quad \text{Area}(\text{ConvexHull}(c))
\]

The total convex hull area is defined as the summation of the areas of all clusters for a given prompt \( p \) at temperature \( t \) without noise points, i.e., \(c \neq -1\)).

\[
A(p, t) = \sum_{c \in \mathbf{L}, c \neq -1} \text{Area}(\text{ConvexHull}(c))
\]

The final result provides the uncertainty level of the VLM's responses to the prompt and temperature setting. In this context, a larger area indicates a higher uncertainty, while a smaller area indicates a lower uncertainty.

\subsection{Analysis of Convex Hull-based Uncertainty Quantification}

The method relies on embedding the model’s responses in a high-dimensional space, clustering these responses, and using the convex hull of the clusters to measure uncertainty.

\subsubsection{Mathematical Justification}

Given a prompt \( p \), let \( R(p) = \{r_1, r_2, \dots, r_n\} \) represent the set of responses generated by the VLM. Each response \( r_i \in R(p) \) is embedded into a \( d \)-dimensional space, resulting in an embedding vector \( E(r_i) \in \mathbb{R}^d \).

The following analysis proves why the convex hull-based approach captures uncertainty effectively.

The convex hull of a set of points is the smallest convex set containing all the points. The geometric properties of the convex hull make it a suitable tool for measuring the diversity (and thus the uncertainty) of responses, since the area (or volume in higher dimensions) of the convex hull reflects the spatial spread of the points.

\textbf{Lemma:} The area of the convex hull \( \text{ConvexHull}(E(R(p))) \) increases as the diversity of the model’s responses increases.

\textbf{Proof:} 
Let \( E(R(p)) = \{E(r_1), E(r_2), \dots, E(r_n)\} \) be the set of response embeddings. As the diversity of the responses increases, the distances between embedding points in \( E(R(p)) \) will also increase. The convex hull is the minimum convex set containing all these points, and its area is proportional to the spatial distribution of the points.

Let \( S \subset \mathbb{R}^d \) be the set of embeddings with larger pairwise distances between points. By the properties of convex sets, the convex hull of a larger set \( S \) will have a greater area than that of a set with points more closely clustered together:
\[
A(\text{ConvexHull}(S_1)) < A(\text{ConvexHull}(S_2)) \quad \text{if}
\]
\[
\|E(r_i) - E(r_j)\| < \|E(r'_i) - E(r'_j)\|
\]
where \( S_1 \) and \( S_2 \) are two sets of embeddings with increasing diversity. Thus, a more diverse set of responses corresponds to a larger convex hull area, indicating greater uncertainty.

To capture the structure of response embeddings, we apply the DBSCAN algorithm, which identifies clusters of similar responses. If the responses generated by the VLM are consistent, the embeddings will form tight clusters with small convex hull areas. In contrast, if the responses are highly uncertain, the clusters will be more dispersed, leading to larger convex hull areas.

\textbf{Lemma:} The total uncertainty for a prompt \( p \) is proportional to the sum of the convex hull areas of all clusters generated by DBSCAN.

\textbf{Proof:} 
Let \( L = \{c_1, c_2, \dots, c_k\} \) be the set of clusters identified by DBSCAN. For each cluster \( c_i \), we compute the convex hull \( \text{ConvexHull}(c_i) \) and its area \( A(c_i) \). The total uncertainty is then:
\[
A(p, t) = \sum_{i=1}^{k} A(\text{ConvexHull}(c_i))
\]
Since the area of each convex hull \( A(\text{ConvexHull}(c_i)) \) reflects the diversity within each cluster, the sum of these areas measures the overall spread of the response embeddings. A larger total area corresponds to more spread out clusters, which indicates greater uncertainty in the model’s responses.

\subsubsection{Temperature Sensitivity and Uncertainty}

The temperature parameter \( t \) affects the stochasticity of the VLM’s outputs. As \( t \) increases, the model produces more diverse and uncertain responses. Formally, for a higher temperature \( t \), the spread of the embedding points increases, leading to larger convex hulls:
\[
\frac{\partial A(p, t)}{\partial t} > 0
\]
This shows that the uncertainty \( A(p, t) \) increases with the temperature, reflecting the model’s sensitivity to the temperature parameter.

The convex hull-based method works for uncertainty quantification because it leverages the geometric properties of the response embeddings. By clustering and measuring the area of the convex hulls, the method captures both the consistency and the diversity of the model’s responses. As the diversity of the responses increases, the convex hull area grows, reflecting higher uncertainty. Therefore, the proposed approach provides a sound theoretical foundation for quantifying uncertainty in VLM outputs.

\subsection{Experimental Results}
In this study, five different cases are conducted to evaluate the uncertainty of the selected VLM's responses at different temperature settings, i.e., 0.001, 0.25, 0.50, 0.75, and 1.00. 30 different radiology reports were generated for each image in the X-ray dataset given the prompt, i.e., "Generate a comprehensive radiology
reports for the entered chest X-ray image."

\subsubsection{Case Study I: A temperature setting of 0.001}
\begin{figure}
    \centering
        \centering
        \includegraphics[scale=0.35]
        {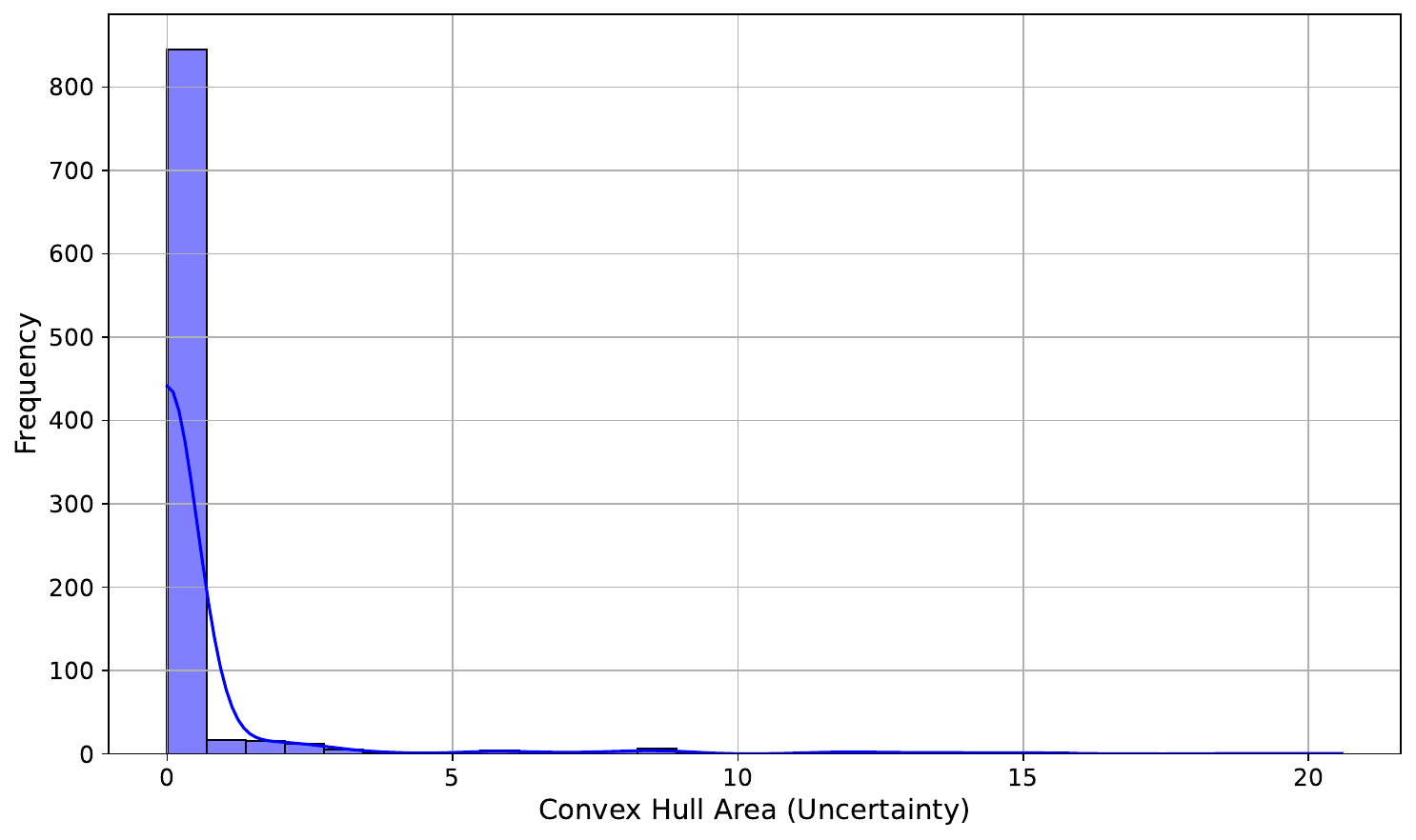}
    \caption{Uncertainty distribution at the temperature setting=0.001}
    \label{fig:unc_dist_0.001}
\end{figure}

Figure \ref{fig:unc_dist_0.001} shows a histogram representing the uncertainty distribution of the convex hull areas in the reports generated from the VLM at a temperature setting of 0.001. The temperature setting is selected to be close to 0, i.e., 0.001, since the temperature value must remain positive. The temperature value can also be set to 0 for tasks requiring more reliable responses but is not ideal for tasks requiring creativity or varied responses. In Figure \ref{fig:unc_dist_0.001} and subsequent histogram figures, the x-axis indicates the convex hull area, while the y-axis denotes the frequency, i.e., the number of occurrences of each convex hull area. 
According to Figure \ref{fig:unc_dist_0.001}, there is almost no degree of diversity in the responses from VLM. The histogram shows a peak at low convex hull areas close to 0, i.e., the selected VLM has a strong tendency to generate confident responses at very low temperature settings. It is also expected that a low temperature setting (close to 0) results in low diversity in responses generated by the model. The figure also shows a small distribution across a range of higher convex hull areas with lower frequencies. This indicates that very few responses fall into the range of higher uncertainty. In general, the histogram figure indicates that most responses generated by VLM are highly certain or low uncertainty at a temperature setting of 0.001.

\begin{figure}[ht!]
    \centering    
    \begin{subfigure}[b]{0.450\linewidth}
        \centering
        \includegraphics[width=\linewidth]{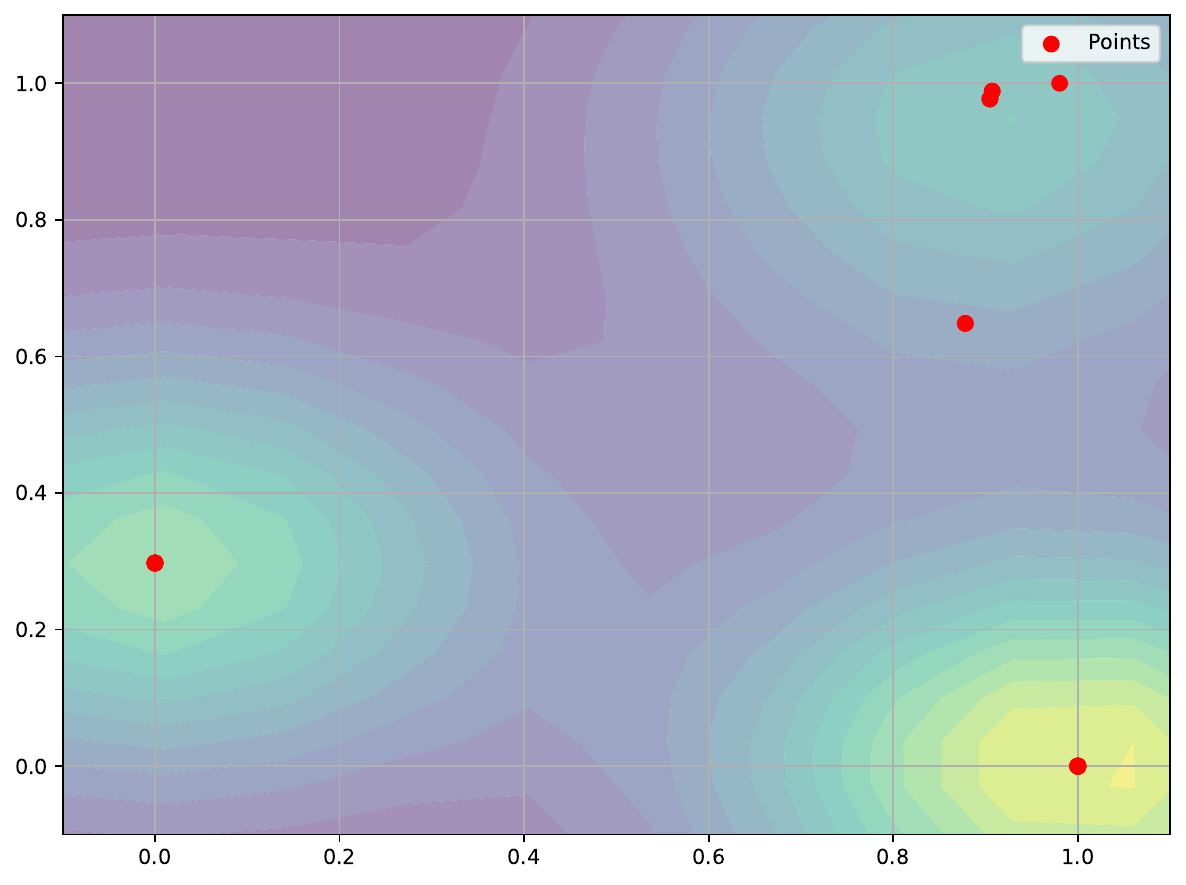}
        \caption{Instance 1}
        \label{fig:least_uncertain_1}
    \end{subfigure}
    \hfill
    \begin{subfigure}[b]{0.45\linewidth}
        \centering
        \includegraphics[width=\linewidth]{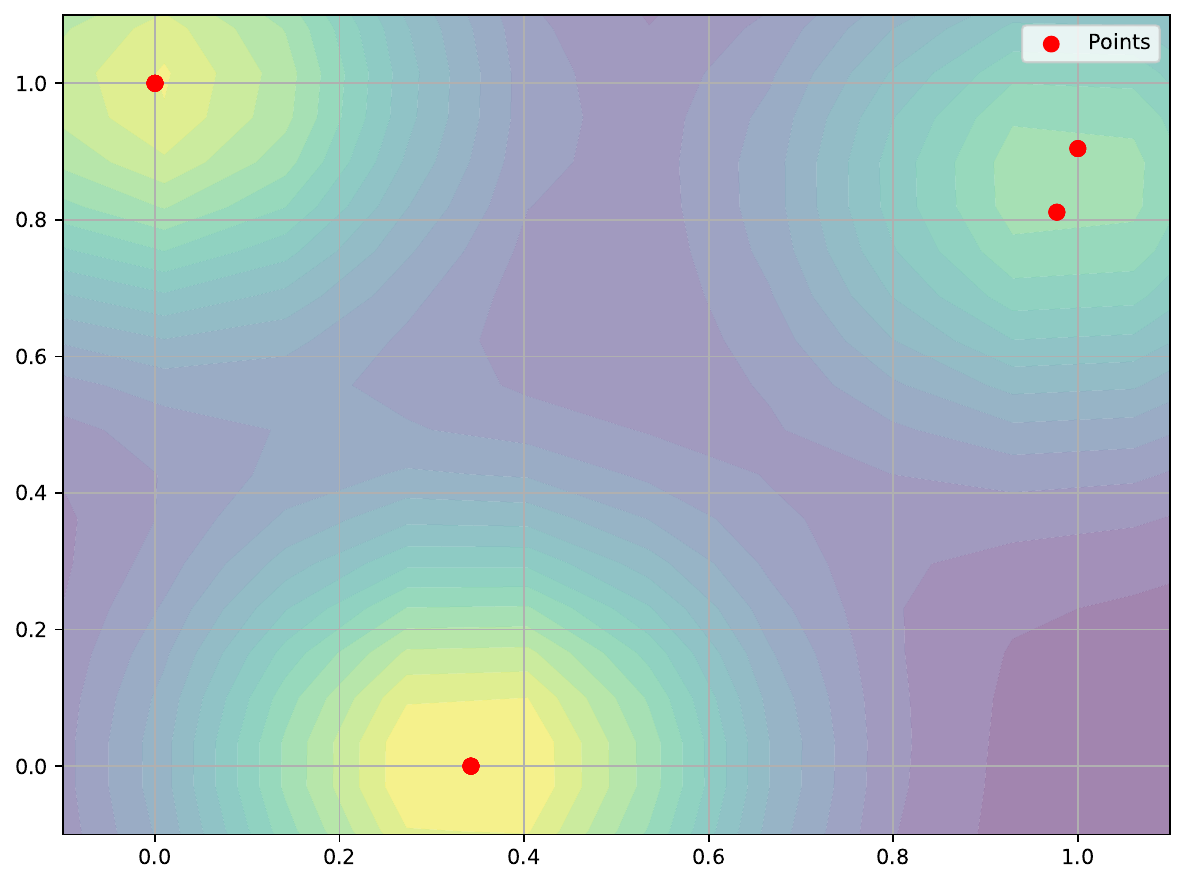}
        \caption{Instance 2}
        \label{fig:least_uncertain_2}
    \end{subfigure}
    \hfill
    \caption{Most uncertain instances (a-b) at the temperature setting of 0.001}
    \label{fig:uncertainty_comparison_0.00}
\end{figure}

Figure \ref{fig:uncertainty_comparison_0.00} shows the two most uncertain instances at a temperature setting of 0.001 based on convex hull areas on the contour maps. Each subfigure represents a 2D visualization corresponding to an embedding of a generated response. The least uncertain instances are also generated on the contour maps; however, they are not given in this study since showing very little or no uncertainty, i.e., the model responses remain consistent and certain under this deterministic setting. In the figure, the red dots represent the data point (generated response in 2D) for each instance, and the background color ranges from purple to turquoise and yellow, with yellow representing areas of higher uncertainty, and turquoise areas indicating lower uncertainty. It is expected to see one or more convex hulls (outlined by a dashed line) evaluate the level of uncertainty. However, in this case, all most uncertain instances show a similar pattern with turquoise and yellow areas near the red dots, but without any cluster. The model generated more consistent and confident responses at a temperature setting of 0.001 as expected. 

\subsubsection{Case Study II - A temperature setting of 0.25} 
\begin{figure}
    \centering
        \centering
        \includegraphics[scale=0.35]{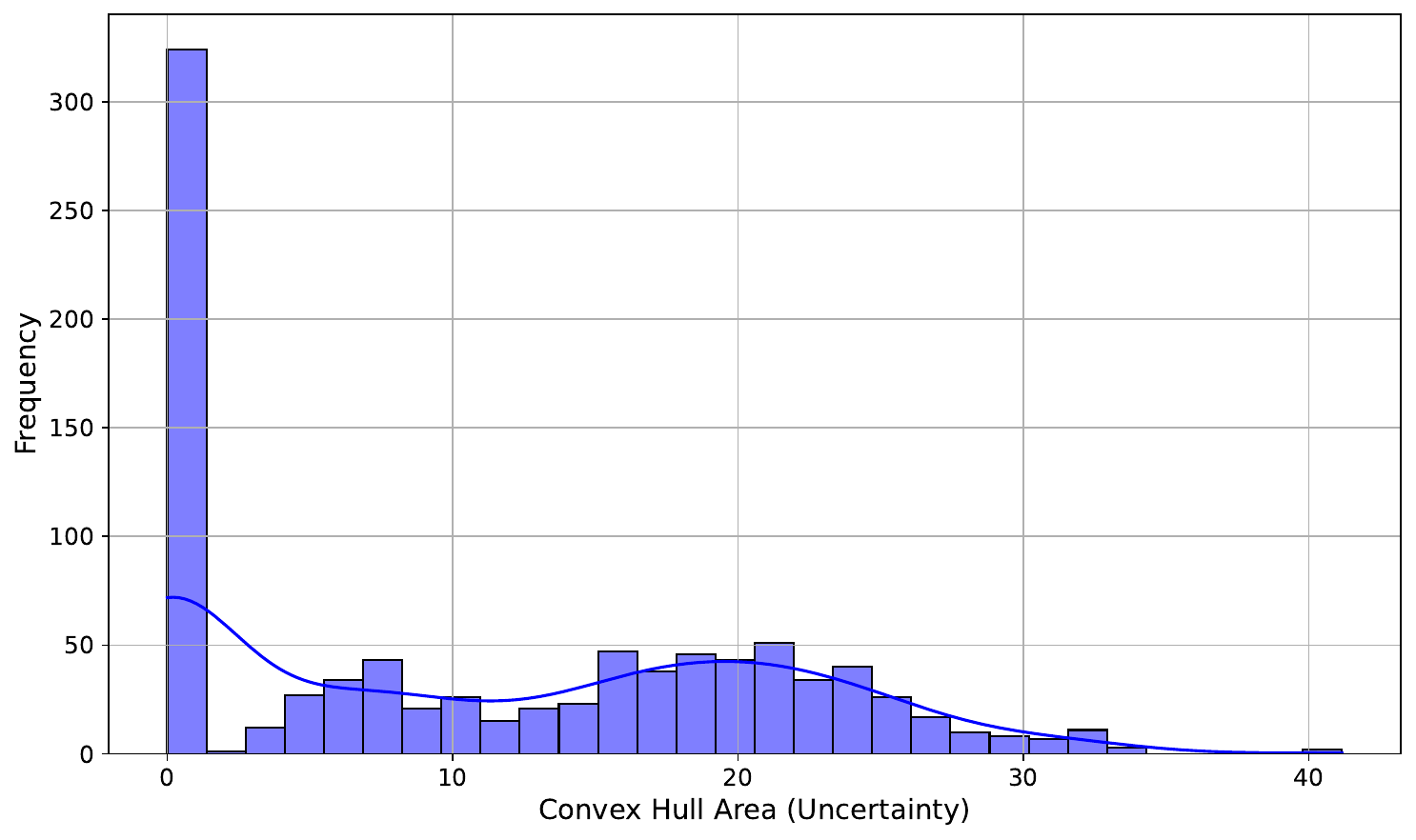}
    \caption{Uncertainty distribution at the temperature setting=0.25}
    \label{fig:unc_dist_0.25}
\end{figure}

Figure \ref{fig:unc_dist_0.25} shows a histogram representing the uncertainty distribution of the convex hull areas in the reports generated from the VLM at a temperature of 0.25. According to the figure, there is some degree of diversity in the VLM's responses, which do not exist at the temperate setting of 0.001. The histogram shows a highly skewed distribution with a significant concentration of responses around a convex hull area close to 0, i.e., a large number of generated reports demonstrate very low uncertainty. In addition, it has a scattered distribution across a range of higher convex hull areas with lower frequencies. This indicates that most of the VLM responses remain highly certain, while the majority of the responses are with low uncertainty. However, the presence of a peak in low convex-hull areas close to 0 indicates that the selected VLM has a strong tendency to generate confident responses.

\begin{figure}[h]
    \centering
    \begin{subfigure}[b]{0.45\linewidth}
        \centering
        \includegraphics[width=\linewidth]{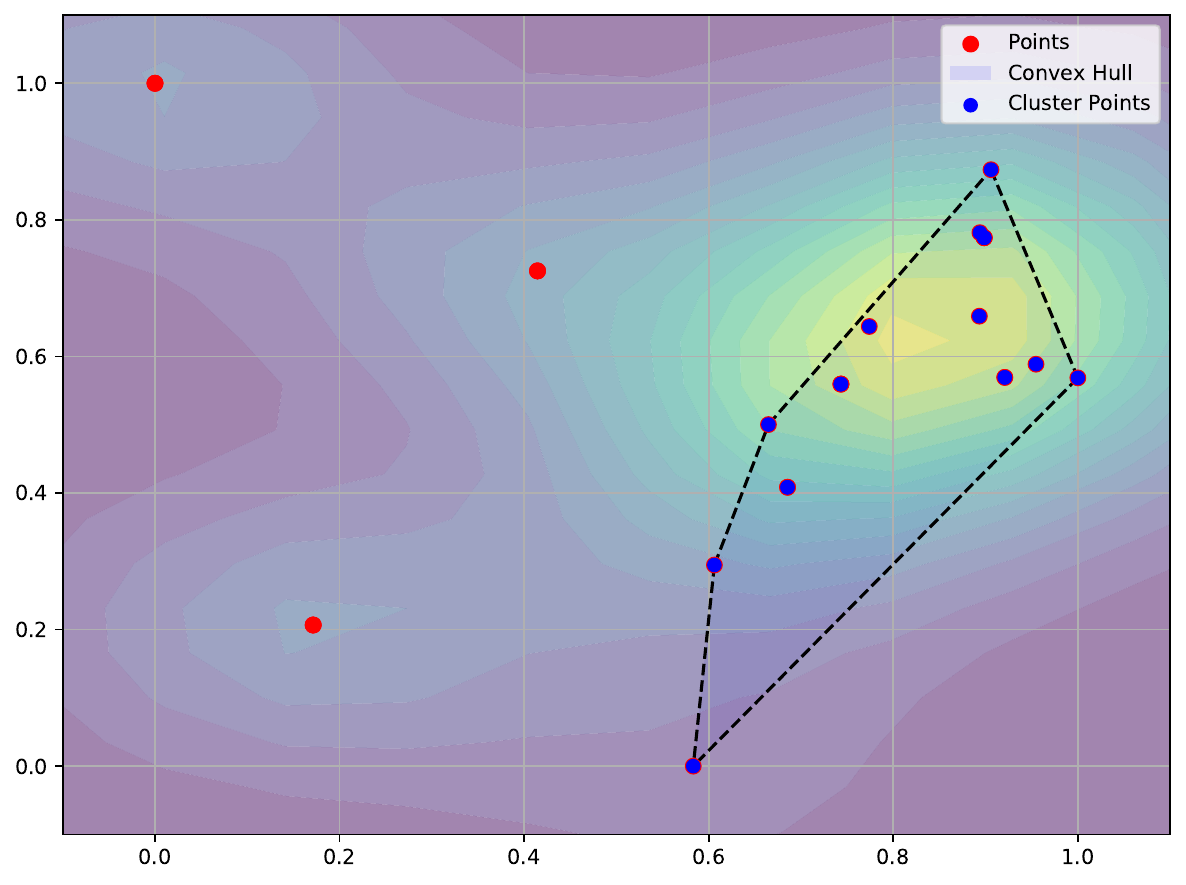}
        \caption{Instance 1}
        \label{fig:least_uncertain_1}
    \end{subfigure}
    \hfill
    \begin{subfigure}[b]{0.45\linewidth}
        \centering
        \includegraphics[width=\linewidth]{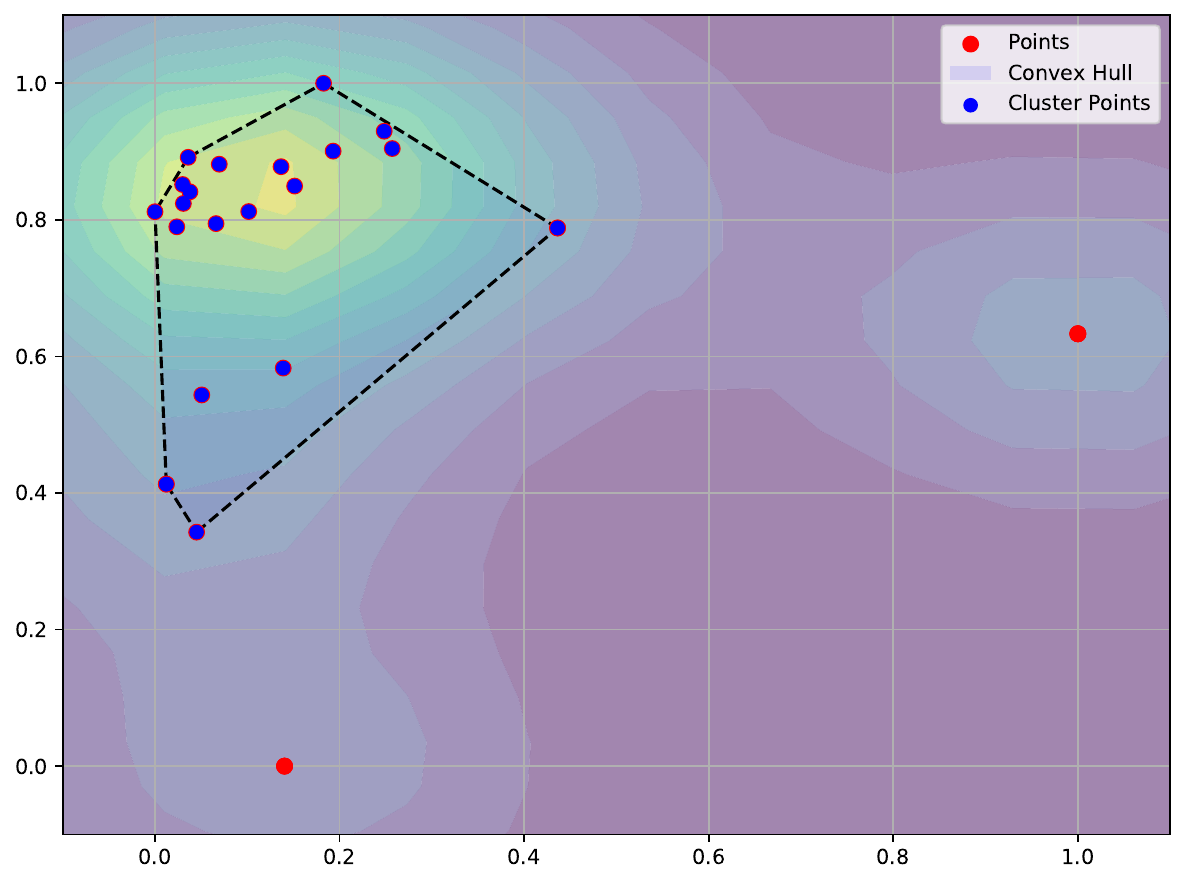}
        \caption{Instance 2}
        \label{fig:least_uncertain_2}
    \end{subfigure}
    \hfill
    \caption{Most uncertain instances (a-b) at the temperature setting of 0.25}
    \label{fig:uncertainty_comparison_0.25}
\end{figure}

Figure \ref{fig:uncertainty_comparison_0.25} illustrates the two most uncertain instances at a temperature setting of 0.25 on the contour maps. The dark dashed forms encapsulate groups of cluster points to identify the convex hull area for the uncertain instances. For these two instances, the convex hull encloses most of the red points and shows a region regarding the diversity in uncertainty. This figure indicates that the model with a temperature setting of 0.25 is less confident in its responses than the one with a setting of 0.001, as anticipated.

\subsubsection{Case Study III: A temperature setting of 0.50}
\begin{figure}
    \centering
        \centering
        \includegraphics[scale=0.35]{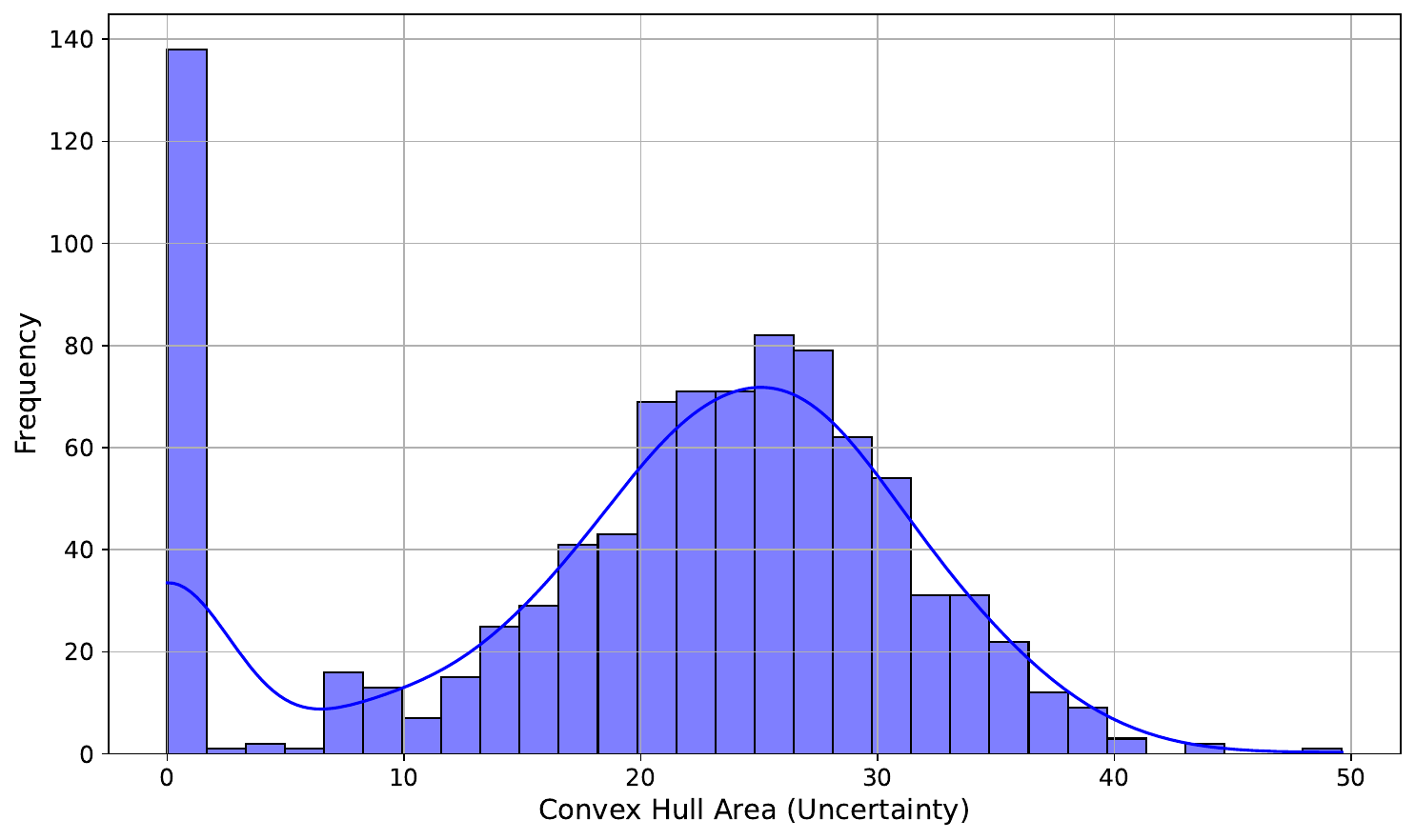}
    \caption{Uncertainty distribution at the temperature setting=0.50}
    \label{fig:unc_dist_0.50}
\end{figure}

Figure \ref{fig:unc_dist_0.50} illustrates a histogram depicting the distribution of the convex hull area values corresponding to the uncertainty of the responses generated from the VLM at a temperature setting of 0.50. In this medium-level temperature setting, the VLM generates responses with a balanced level of diversity. The histogram displays two distinct patterns, i.e., a sharp peak at a convex hull area close to 0, and a norma distribution centered around a convex hull area of 25. The sharp peak close to 0 indicates that a significant number of generated responses have very low uncertainty with high confidence, while the normal distribution curve indicates a wide range of responses with moderate uncertainty. 

\begin{figure}[h]
    \centering
    \begin{subfigure}[b]{0.45\linewidth}
        \centering
        \includegraphics[width=\linewidth]{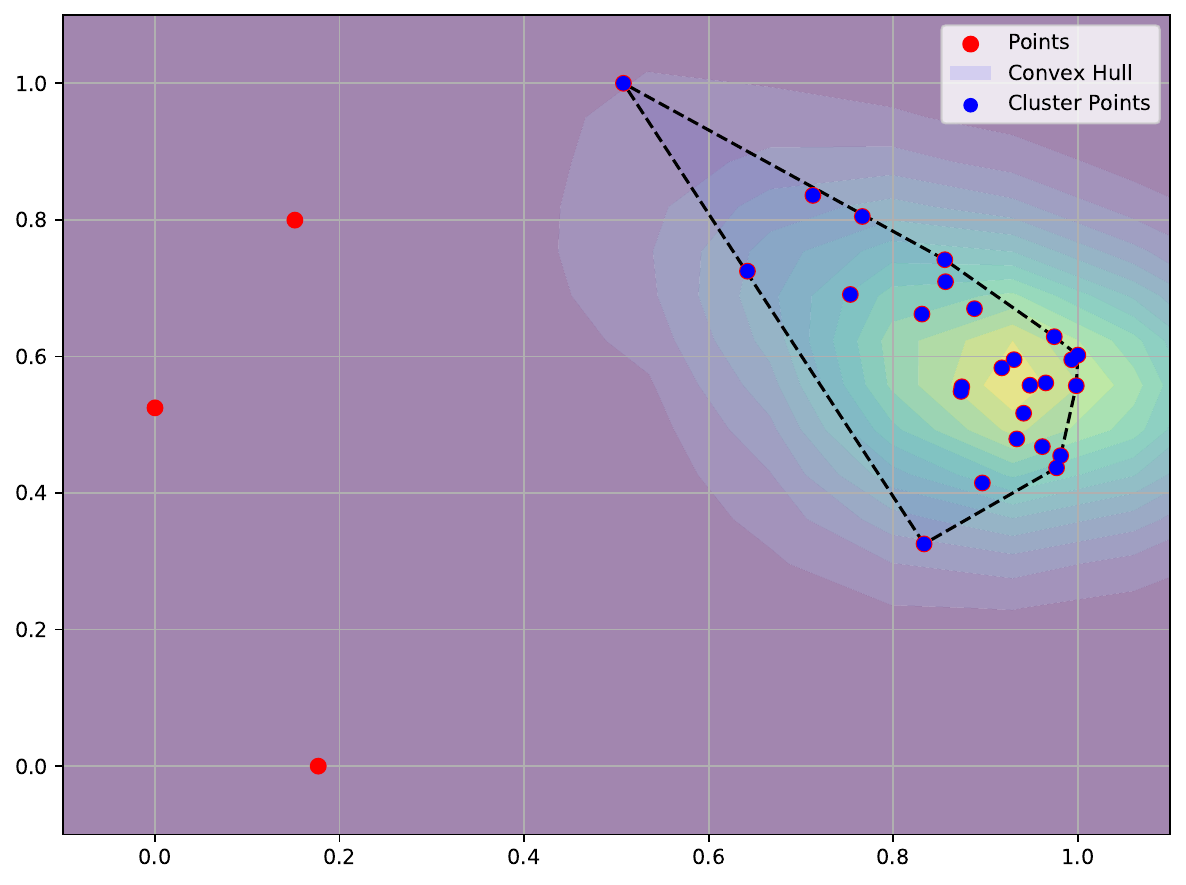}
        \caption{Instance 1}
        \label{fig:least_uncertain_1}
    \end{subfigure}
    \hfill
    \begin{subfigure}[b]{0.45\linewidth}
        \centering
        \includegraphics[width=\linewidth]{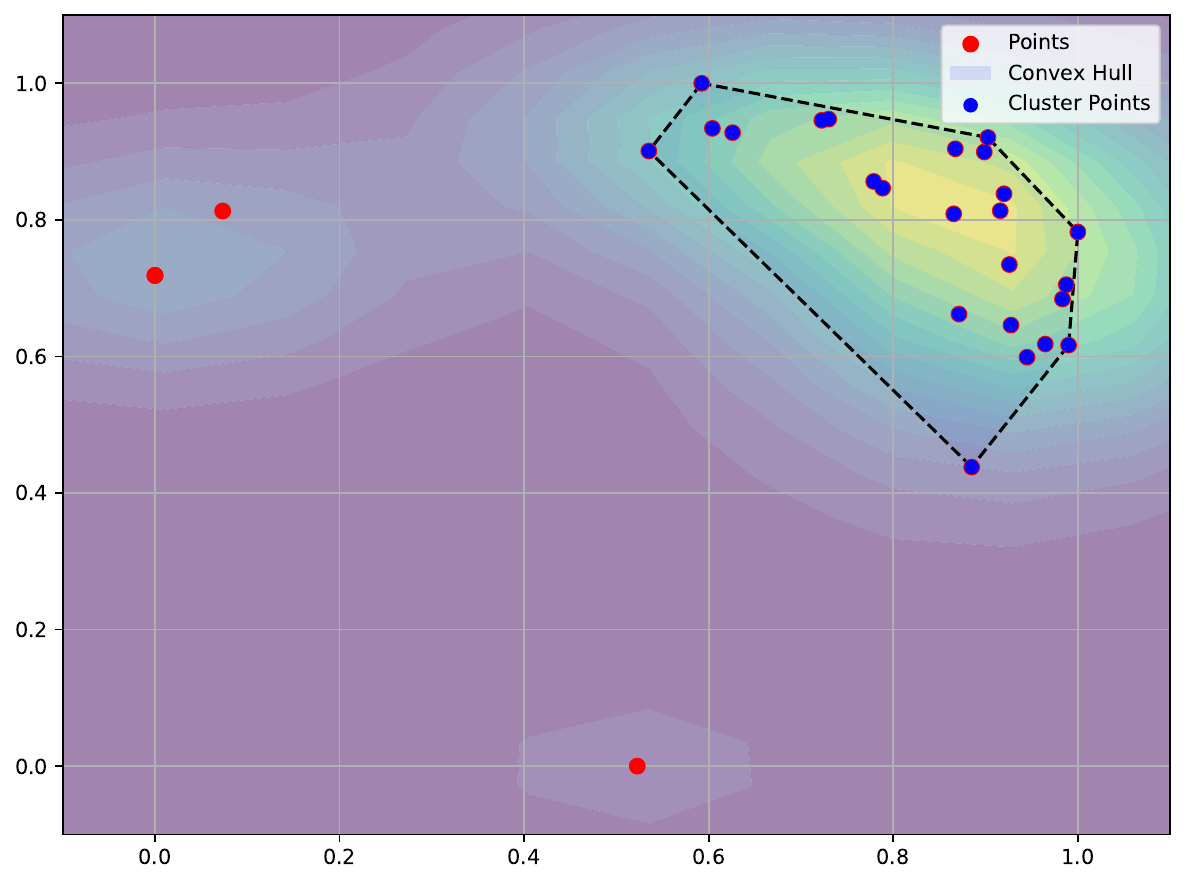}
        \caption{Instance 2}
        \label{fig:least_uncertain_2}
    \end{subfigure}
    \hfill
    \caption{Most uncertain instances at the temperature setting=0.50}
    \label{fig:uncertainty_comparison_0.50}
\end{figure}

Figure \ref{fig:uncertainty_comparison_0.50} depicts the two most uncertain instances at a temperature setting of 0.50 on a contour map. The figure consists of two plots with the background colored contours ranging from purple (lower uncertainty) to turquoise and yellow (higher uncertainty) to emphasize the degree of confidence. These two instances have dark dashed forms that enclose groups of points, outlining the convex hull to show the area where the uncertainty is highest. As the temperature setting is increased, the model tends to be less confident and more uncertain in each instance, as indicated by the presence of larger convex hulls and more significant yellow areas.

\subsubsection{Case Study IV: A temperature setting of 0.75}
\begin{figure}
    \centering
        \centering
        \includegraphics[scale=0.35]{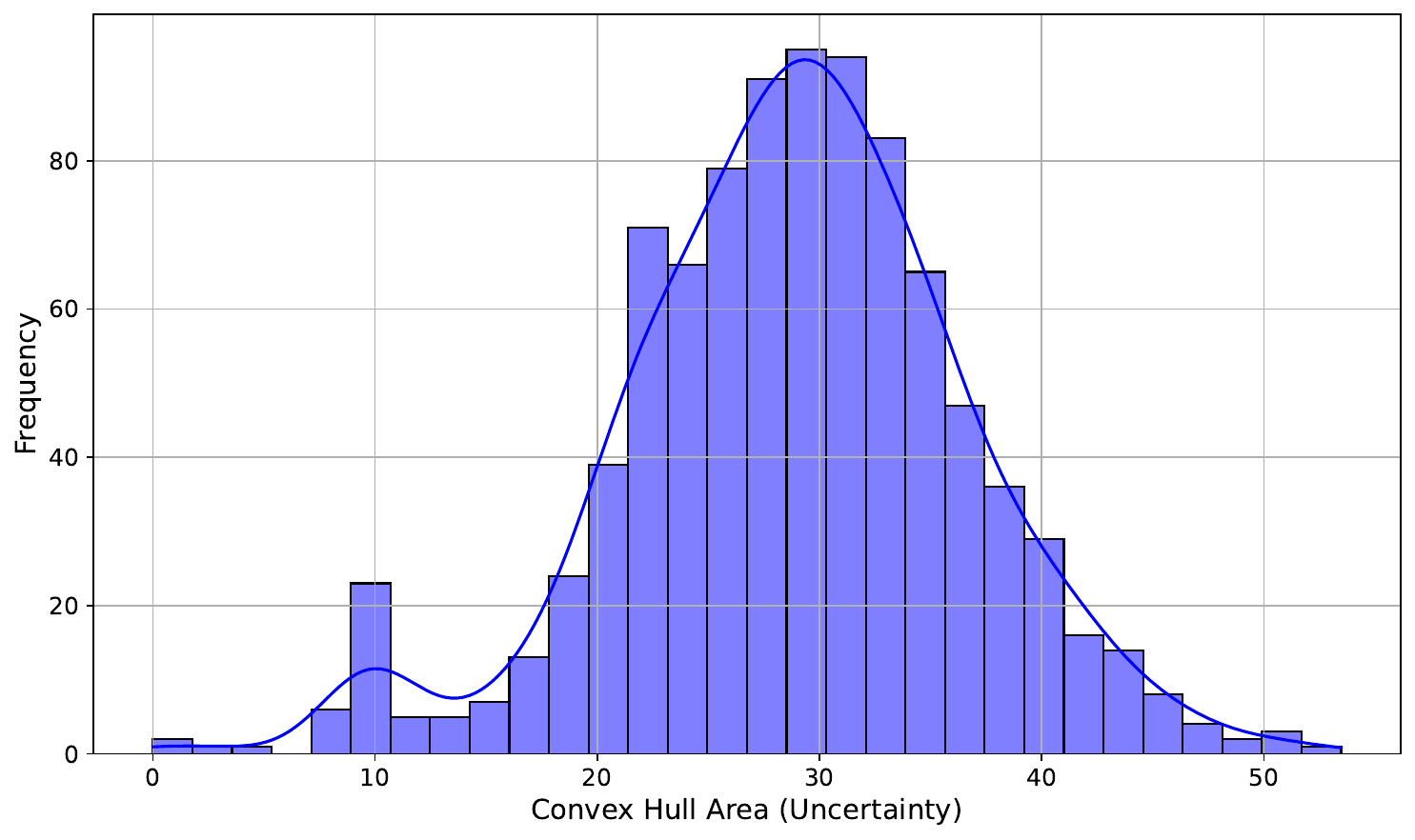}
    \caption{Uncertainty distribution at the temperature setting=0.75}
    \label{fig:unc_dist_0.75}
\end{figure}

Figure \ref{fig:unc_dist_0.75} provides a histogram showing the uncertainty distribution of the responses generated by VLM based on the area of the convex hull at a temperature setting of 0.75. The histogram reveals a bimodal distribution with two peaks, i.e., the main peak is around a convex hull area of about 30, while a smaller peak is present about 10. The main peak indicates that the most common convex hull area (uncertainty) is around 30. Compared to previous cases, it does not have a sharp peak near 0. In other words, it indicates that most generated responses have a moderate level of uncertainty and fairly consistent reports with controlled uncertainty when the temperature is set to 0.75.

\begin{figure}[h]
    \centering

    \begin{subfigure}[b]{0.45\linewidth}
        \centering
        \includegraphics[width=\linewidth]{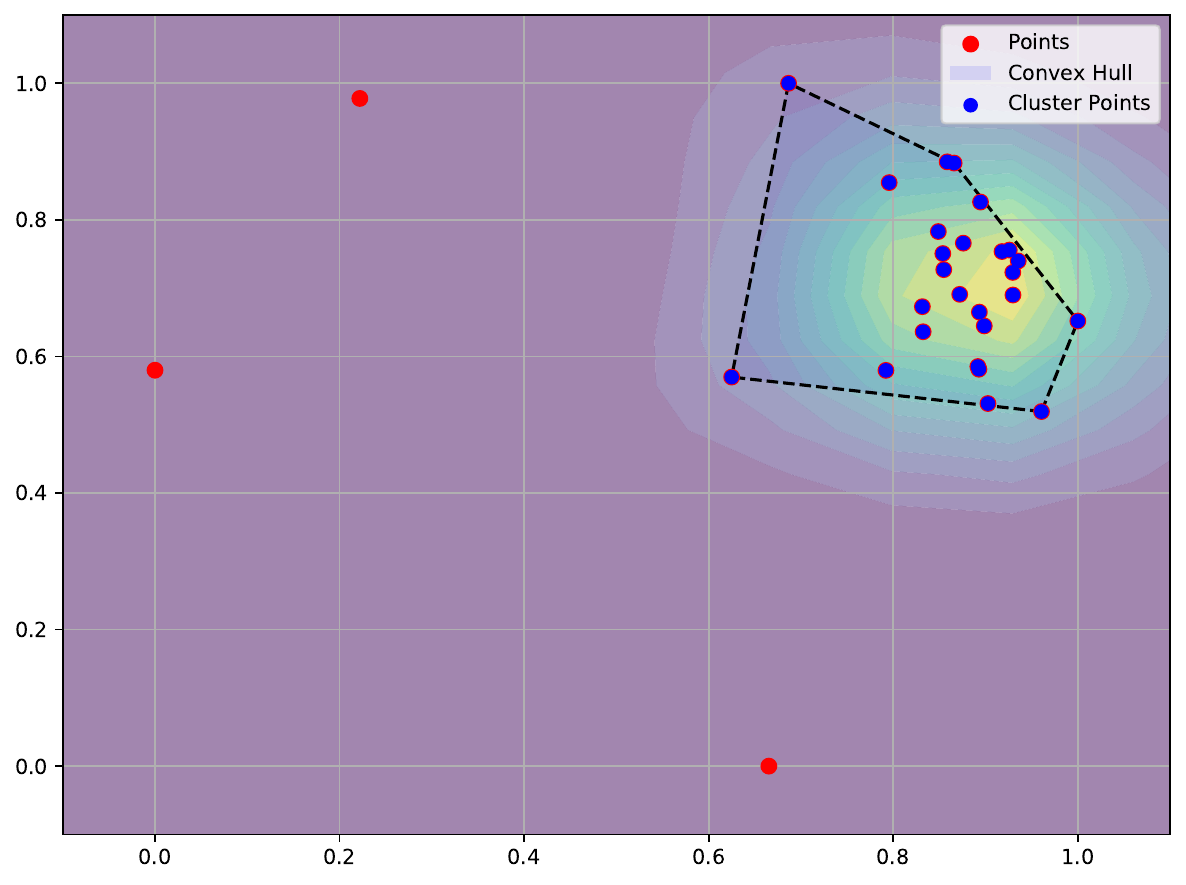}
        \caption{Instance 1}
        \label{fig:least_uncertain_1}
    \end{subfigure}
    \hfill
    \begin{subfigure}[b]{0.45\linewidth}
        \centering
        \includegraphics[width=\linewidth]{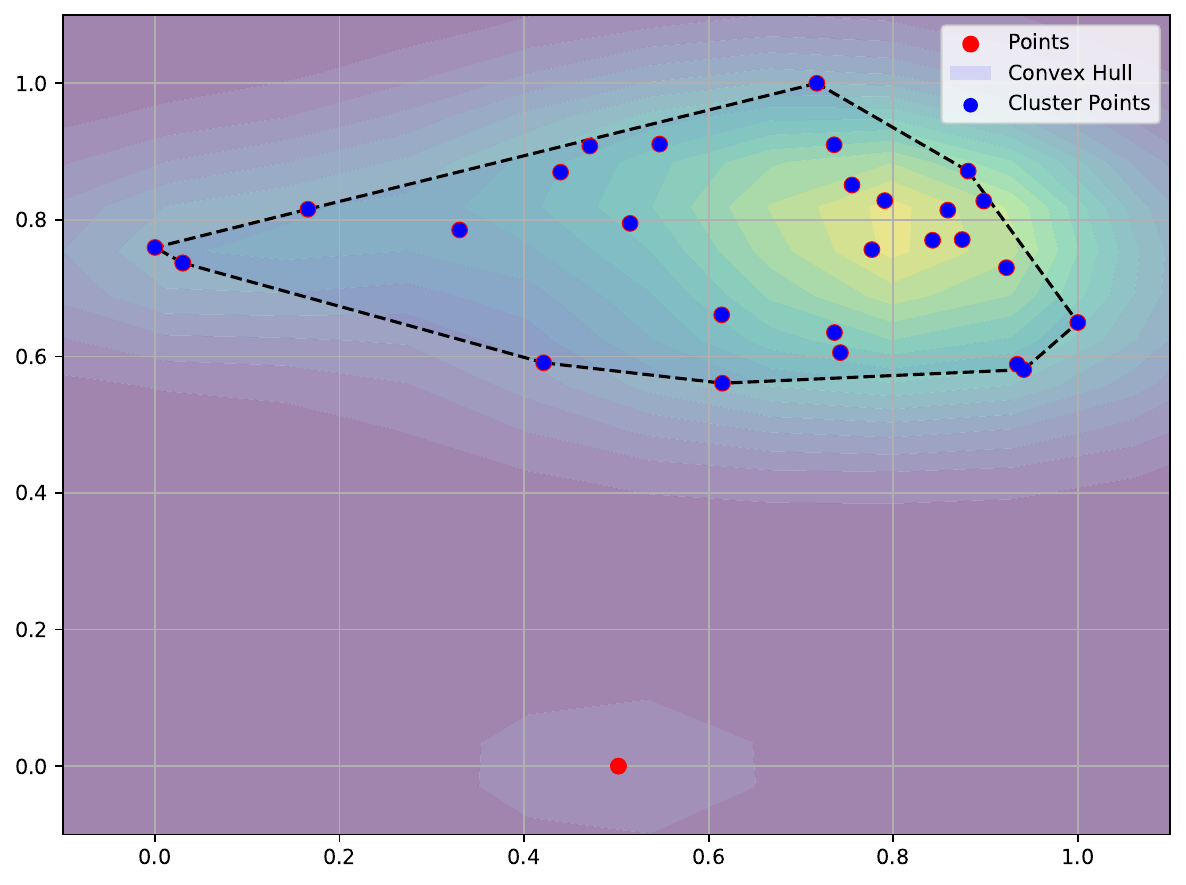}
        \caption{Instance 2}
        \label{fig:least_uncertain_2}
    \end{subfigure}
    \hfill
    \caption{Most uncertain instances at the temperature setting=0.75}
    \label{fig:uncertainty_comparison_0.75}
\end{figure}

Figure \ref{fig:uncertainty_comparison_0.75} shows the most uncertain instances (a-b) for a temperature setting of 0.75 through a contour map. The temperature setting of 0.75 introduces more diversity and uncertainty regarding the VLM responses. Each instance shows different instances of uncertainty, highlighting areas where the model has low confidence in its responses. As expected, at a higher temperature, the model shows a higher uncertainty, as seen in two instances.

\subsubsection{Case Study V: A temperature setting of 1.00}
\begin{figure}
    \centering
        \centering
        \includegraphics[scale=0.35]
        {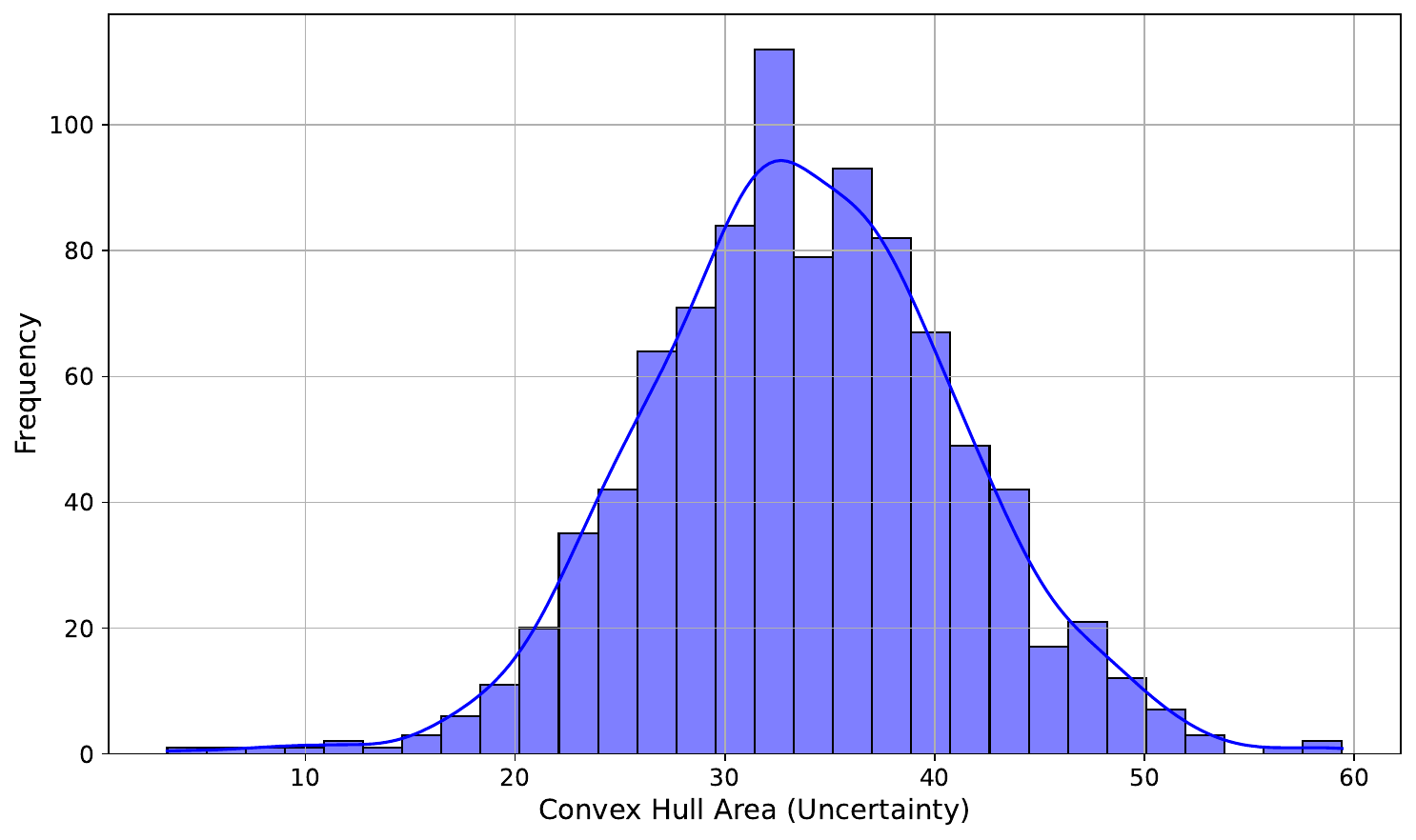}
    \caption{Uncertainty distribution at a temperature setting=1.00}
    \label{fig:unc_dist_1.0}
\end{figure}

Figure \ref{fig:unc_dist_1.0} presents a histogram illustrating the distribution of the convex hull area for the responses generated from VLM when the temperature is set to 1.00.
The histogram reveals a normal distribution and one pattern compared to the previous cases, i.e., a bell-shaped, relatively symmetrical distribution. In other words, it has no sharp peak close to 0 and no peak at low convex hull areas, i.e., neither very low uncertainty nor high confidence. It also indicates that most generated responses have a moderate or high level of uncertainty when the temperature is set to 1.00. According to the histogram, the VLM generates fairly consistent reports with controlled uncertainty under a very high temperature setting. It also highlights the uncertainty concerns in VLM's responses.

\begin{figure}[h]
    \centering

    \begin{subfigure}[b]{0.45\linewidth}
        \centering
        \includegraphics[width=\linewidth]{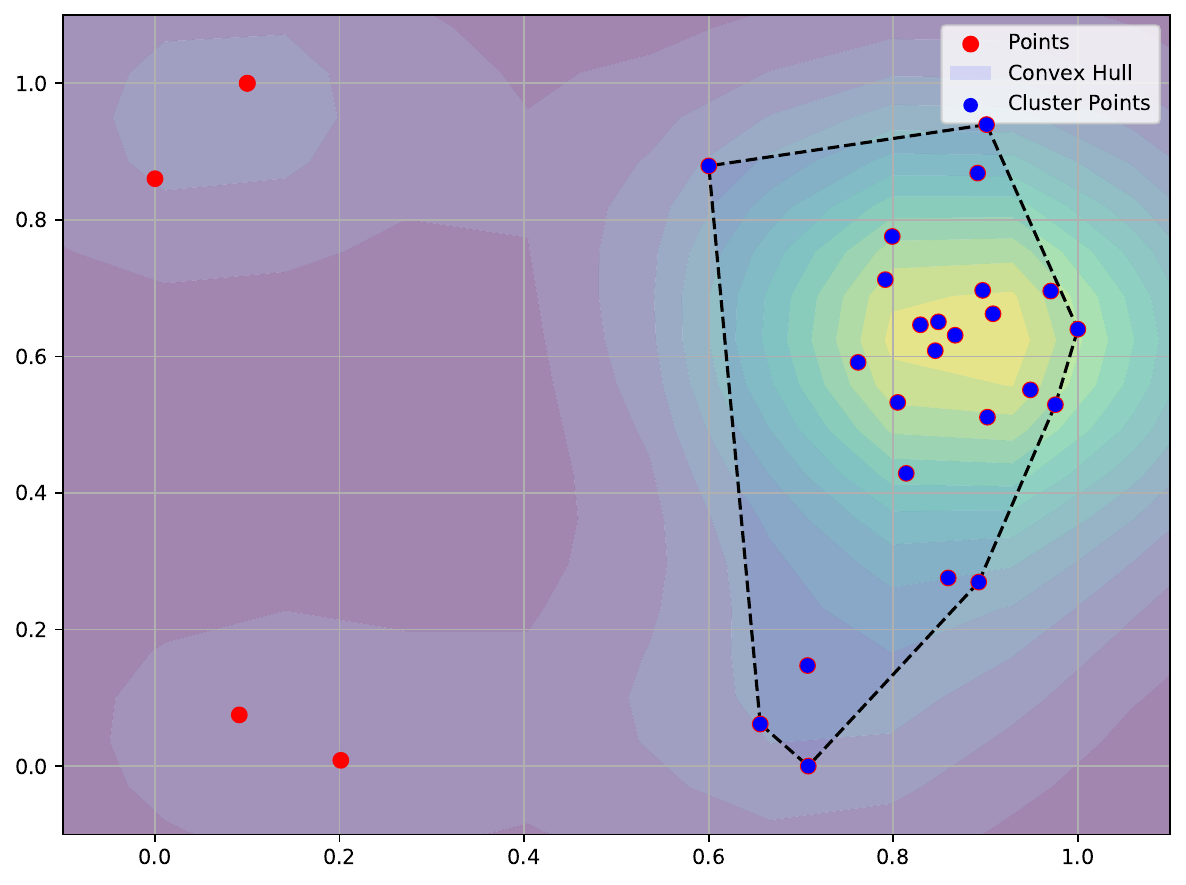}
        \caption{Instance 1}
        \label{fig:least_uncertain_1}
    \end{subfigure}
    \hfill
    \begin{subfigure}[b]{0.45\linewidth}
        \centering
        \includegraphics[width=\linewidth]{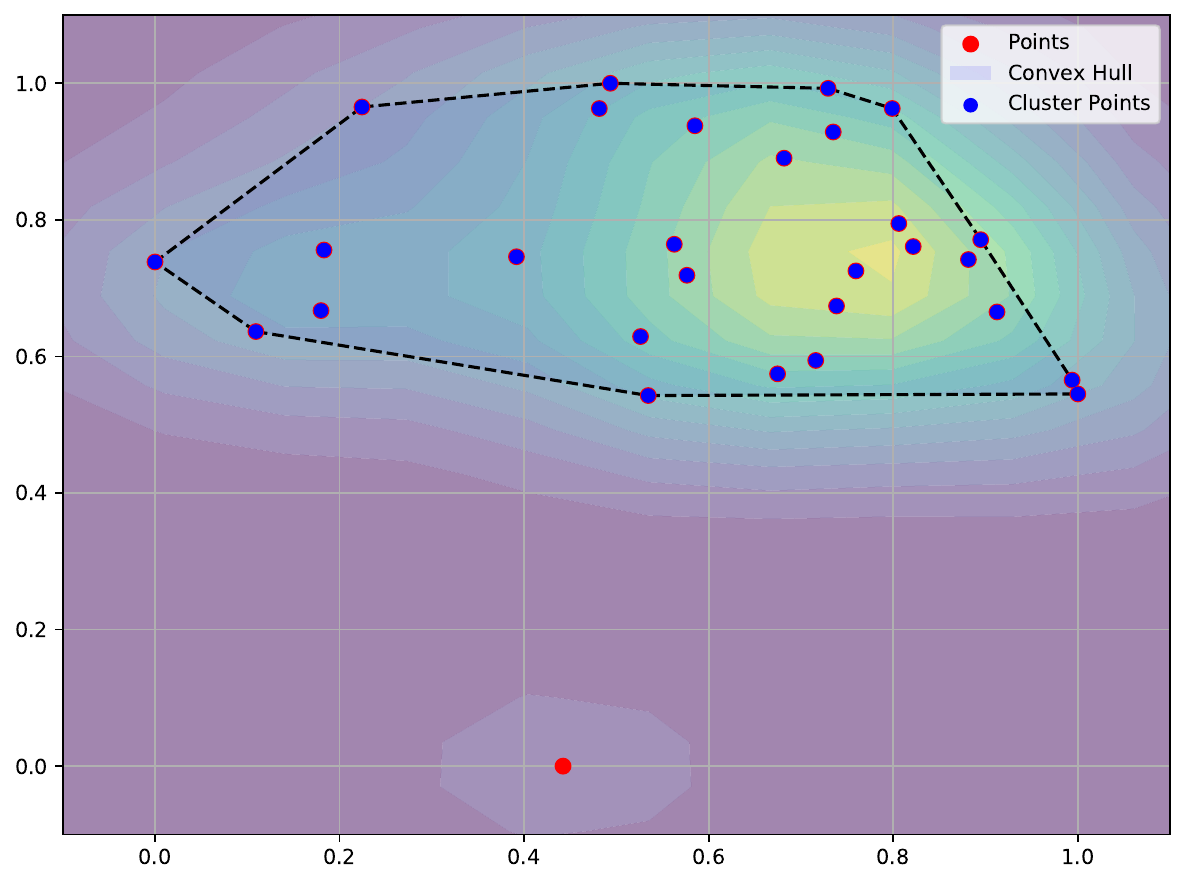}
        \caption{Instance 2}
        \label{fig:least_uncertain_2}
    \end{subfigure}
    \hfill
    \caption{Most uncertain instances at the temperature setting=1.00}
    \label{fig:uncertainty_comparison_1.00}
\end{figure}

Figure \ref{fig:uncertainty_comparison_1.00} shows a significant increase in uncertainty due to leading to less confidence in responses at higher temperature settings. In the figure, each plot visualizes one instance with its associated uncertainty, outlined by a convex hull, as in the previous figures. For two instances, the model tends to moderate and high uncertainty, as antipicated. 

\section{Discussion and Observations}
The results provide valuable insights into the uncertainty of VLM's responses at different temperature settings, i.e., 0.001, 0.25, 0.50, 0.75, and 1.00. The temperature parameter plays a crucial role in the diversity of the VLM's resonances, directly affects the uncertainty. Observations are given for each case below.

\begin{itemize}
    \item At a temperature setting of 0.001: Most responses from the VLM have low uncertainty with highly confident, with minimal diversity. The histogram shows a peak in the lower convex hull areas, i.e., close to 0. The most uncertain instances also show minimal convex hull areas and consistent responses.
    \item At a temperature setting of 0.25: The histogram shows a skewed distribution, with a significant peak at low uncertainty (close to 0) and a scattered distribution across a range of higher convex hull areas at high uncertainty. This indicates that the model generates responses with confident responses, as well as diverse responses with varying levels of uncertainty. 
    \item At a temperature setting of 0.50: The VLM shows a high uncertainty in its responses with two distinct patterns, i.e., a sharp peak close to 0, and a normal distribution centered around a convex hull area of 25. This indicates that the model can generate responses with high confidence while generating responses with high uncertainty. 
    \item At a temperature of 0.75: The results show a bimodal distribution with a main peak at a convex hull area of 30 and a smaller peak around 10. Unlike lower temperature settings, there is no sharp peak near 0, indicating that most responses demonstrate moderate or high uncertainty. 
    \item At a temperature setting of 1.00: VLM's responses tend to a high uncertainty. The histogram reveals a normal distribution for convex hull areas with respect to the uncertainty of VLM-generated responses, i.e., no sharp peak near 0. The normal distribution without sharp peak near 0 indicates higher diversity and uncertainty in the model responses. 
    \item The most uncertain instances indicate the importance and uncertainty concerns of VLM's responses at high-temperature settings. VLM generates diverse responses and leads to higher uncertainty at high temperature settings. 
    
\end{itemize}

In addition to the results given in figures, Table \ref{tab:tab1} provides detailed statistical results, i.e., mean, standard deviation (Std), minimum (Min), maximum (Max) and cumulative averages within a certain percentage of the results (10\%, 25\%, 50\%, 75\%, 90\%), regarding the evaluation of uncertainty in responses generated by VLM for temperature settings. These results provide a detailed view of the uncertainty in responses from VLM.
\begin{itemize}
    \item Mean: The average values increase steadily from 0.001 at the temperature setting of 0.001 to 0.3117 at the temperature setting of 1.00, indicating a trend that increases as the temperature increases. The difference between the lowest and highest mean values is 3115 times. As anticipated, the increase in temperature leads to greater diversity in responses, contributing to higher overall uncertainty in the model's responses. 

    \item The standard deviation (Std): It represents the variability or spread in the responses, with values ranging from 0.0012 at the temperature setting of 0.001 to 0.1441 at the temperature setting of 1.00. The relatively high standard deviation at intermediate temperatures (e.g., 0.1680 at the temperature setting of 0.50) indicates greater diversity at these settings. The difference between the lowest and highest Std is 114 times.

    \item Minimum (Min) and Maximum (Max): Min values are \textit{0.0000} for all temperature settings. This means that the model can also provide confident responses at high temperature settings. On the other hand, maximum (Max) values vary from 0.0232 at the temperature setting of 0.001 to 0.7257 at the temperature setting of 1.00. The diffrence between the lowest and highest max is 31 times.

    \item Cumulative Averages: The percentage-based values, i.e., 10\%, 25\%, 50\%, 75\%, and 90\%, illustrate the cumulative averages of the uncertainty values within the selected percentage. In the table, the 10\% refers to the average value of the lowest 10\% of uncertainty values while 90\%  the average of the lowest 90\% of that. These cumulative averages reveal how uncertainty behaves across different portions of the dataset. It increases along with a high percentage and temperature setting. 

\end{itemize}

\begin{table}[ht]
\centering
\caption{The statistical results of convex hull-based uncertainty evaluation for temperature settings}\label{tab:tab1}
\begin{tabular}{lcccccc}
\hline
\textbf{Temp.} &  \textbf{0.001} & \textbf{0.25} & \textbf{0.50} & \textbf{0.75} & \textbf{1.00} & \textbf{0.001-1.00} \\
\hline
Mean & 0.0001 & 0.1188 & 0.1698 & 0.2473 & 0.3117 & 3115 * \\
Std  & 0.0012 & 0.1725 & 0.1680 & 0.1734 & 0.1441 & 114 *\\
Min  & 0.0000 & 0.0000 & 0.0000 & 0.0000 & 0.0000 & N/A \\
Max  & 0.0232 & 0.7202 & 0.7422 & 0.7831 & 0.7257 & 31 *\\
10\% & 0.0000 & 0.0000 & 0.0000 & 0.0034 & 0.1307 & N/A \\
25\% & 0.0000 & 0.0000 & 0.0000 & 0.1019 & 0.2123 & N/A \\
50\% & 0.0000 & 0.0226 & 0.1390 & 0.2359 & 0.3019 & N/A \\
75\% & 0.0000 & 0.1952 & 0.2870 & 0.3698 & 0.4073 & N/A \\
90\% & 0.0000 & 0.3836 & 0.4150 & 0.4908 & 0.5131 & N/A \\
\hline
\end{tabular}
\end{table}


Additionally, there is a critical need to improve the trustworthiness of AI in VLMs from data preparation to model evaluation. The appendix provides several examples of X-ray images along with responses (i.e., radiology reports) generated by the model. Although the used dataset is publicly available, it includes several noisy or irrelevant images; see Appendices B, D, F, G and H. The model generates reasonable radiology reports for noisy or irrelevant images, which should not occur. This indicates the importance of data preprocessing that ensures that datasets include high-quality images to improve the uncertainty in responses from the model. Furthermore, the integration of explainable AI (XAI) methods into VLMs can be considered to provide explainability and transparency with regard to VLMs' responses. Improving data quality and integrating explainable AI into VLMs can significantly increase overall model performance in terms of uncertainty.


\section{Conclusion}
This study proposes a convex hull-based approach to quantifying uncertainty in Vision-Language Models (VLMs) applied to generating radiology reports. In this study, the LLM-CXR is selected as the VLM, and radiology reports are generated from chest X-ray images for the given prompt at various temperature settings (0.001, 0.25, 0.50, 0.75, and 1.00). The experimental results indicated that uncertainty is still a serious concern as a result of the nature of VLMs and can be significantly higher for high-temperature settings. The proposed approach provides a key metric for developing more reliable VLMs and allows for the improved evaluation of VLMs' responses.  
  %
%
Furthermore, future work could explore the impact of a given prompt and varying temperature settings on the level of uncertainty in VLM's responses to better manage uncertainty.

\bibliographystyle{IEEEtran}
\bibliography{references}

\newpage
\appendix
\subsection{Least Uncertain Instance at the Temperature
Setting of 0.001}
\begin{figure}[h]
    \centering
    \begin{minipage}{1.0\linewidth}
        \centering
        \includegraphics[width=8cm,height=12cm,keepaspectratio]{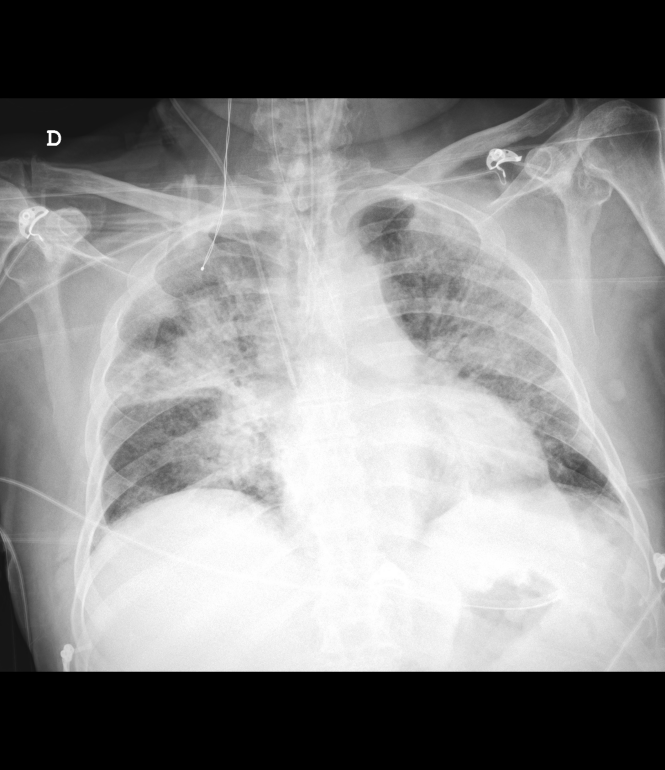}
        \caption*{X-Ray Image}
    \end{minipage}%
    \hfill
    \begin{minipage}{1.0\linewidth}
        \centering
\begin{Verbatim}
Image: covid-19-pneumonia-22-day2-pa.png

Responses:
-- covid-19-pneumonia-22-day2-pa.png
-- No acute intrathoracic process. 
-- No acute intrathoracic process. 
-- No acute cardiopulmonary process. 
-- No acute cardiopulmonary process. 
-- No acute intrathoracic process. 
-- No acute intrathoracic process. 
-- No acute intrathoracic abnormality. 
-- Normal chest x-ray. 
-- Normal chest x-ray. 
-- Normal chest radiographs. 
-- No acute intrathoracic process. 
-- No acute intrathoracic process. 
-- No evidence of pneumonia. 
-- No acute cardiothoracic process. 
-- No acute cardiopulmonary process. 
-- No acute intrathoracic process. 
-- No acute intrathoracic process. 
-- No acute cardiothoracic process. 
-- No acute cardiopulmonary process. 
-- No acute cardiothoracic process. 
-- Normal chest radiograph. 
-- No acute intrathoracic process. 
-- No acute cardiothoracic process. 
-- No acute cardiopulmonary process. 
-- No evidence of pneumonia. 
-- Normal chest radiographs. 
-- No acute cardiopulmonary process. 
-- No acute cardiopulmonary process. 
\end{Verbatim}
    \end{minipage}
\end{figure}

\newpage
\subsection{Most Uncertain Instance at the Temperature
Setting of 0.001}

\begin{figure}[h]
    \centering
    \begin{minipage}{1.0\linewidth}
        \centering
        \includegraphics[width=8cm,height=9cm]{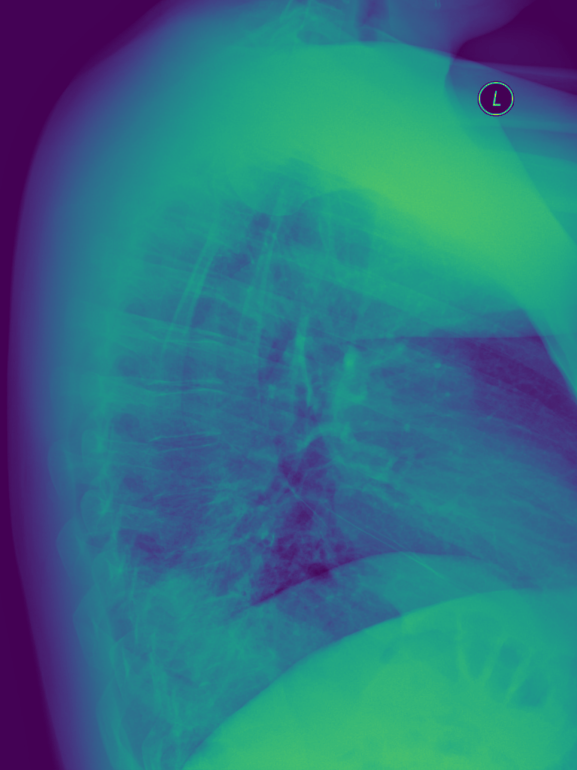}
        \caption*{X-Ray Image}
    \end{minipage}%
    \hfill
    \begin{minipage}{1.0\linewidth}
        \centering
\begin{Verbatim}
Image: d2cdf41a662113279d2ec21af3a4e2_jumbo.jpg

Responses:
-- d2cdf41a662113279d2ec21af3a4e2_jumbo.jpg
-- Findings consistent with pneumonia in the right upper lobe.  Followup radiographs in ___ weeks after treatment are recommended to confirm resolution.   
-- Multifocal pneumonia. 
-- 1.  Increased interstitial markings within the lung bases and at the right lung apex could reflect underlying chronic interstitial lung disease.  However, chronicity is difficult to further evaluate without prior radiographs.  No acute cardiopulmonary process. 2.  Wedge deformity of the mid thoracic spine, age indeterminate.   
-- No acute cardiopulmonary process. 
-- Diffuse interstitial opacities may represent interstitial pulmonary edema.   
-- Findings suggestive of chronic interstitial lung disease bilaterally.  There is no evidence of acute cardiopulmonary process.  Please refer to subsequently obtained CT for further details.   
-- Hilar pulmonary vascular engorgement.  Streaky left basilar airspace opacity is nonspecific but may reflect atelectasis.  Pneumonia in the correct clinical setting is not excluded.   
-- Increased interstitial markings throughout the lungs compatible with chronic lung disease.  No superimposed acute cardiopulmonary process.   
-- Findings consistent with pneumonia in the right lower lobe. 
-- Extensive peribronchial inflammation in both the lingula and right middle lobe.   
-- Findings suggestive of pneumonia in the right mid and lower lung.  Findings were conveyed by Dr. ___ to ___, NP at 10:15 a.m. on ___, upon discovery.   
-- Severe emphysema without focal airspace consolidation to suggest pneumonia.   
-- Multifocal pneumonia. Recommend followup to resolution. 
-- Findings suggesting the presence of pneumonia in the appropriate clinical setting.   
-- Patchy ill-defined opacities within both lung bases, more so on the right, concerning for pneumonia, possibly in the correct clinical setting.  Small bilateral pleural effusions.   
-- Right lower lobe infiltrate compatible with pneumonia in the proper clinical setting.  Recommend repeat after treatment to document resolution.   
-- Increased interstitial markings bilaterally 
-- Diffuse airspace opacities in bilateral lung bases could reflect aspiration or pneumonia. Possible trace right pleural effusion.   
-- 1. Mild diffuse interstitial prominence. 2. No focal consolidation to suggest pneumonia.   
-- 1.  Diffuse, increased interstitial markings in the lung bases could be due to interstitial edema or atypical infection not excluded. 2.  No lobar consolidation.   
-- Findings compatible with right middle lobe pneumonia. 
-- Heterogeneous right lower lobe opacity concerning for pneumonia.  Followup radiographs after treatment are recommended to ensure resolution of this finding.   
-- In comparison with the study of ___, there is little overall change. Again there is extensive hyperexpansion of the lungs with flattening hemidiaphragms consistent with chronic pulmonary disease. There is suggestion of some coarseness of interstitial markings, which could be a chronic interstitial process. However, in the appropriate clinical setting, this would be consistent with some elevated pulmonary venous pressure. Mild hyperexpansion of the lungs, but no evidence of acute focal pneumonia.   
-- Multifocal pneumonia. 
-- Findings suggestive of pneumonia in the right lower lobe.  Recommend repeat after treatment to document resolution.   
-- Right lower lobe pneumonia. 
-- Findings concerning for pneumonia in the right middle and right lower lobe.  Recommend followup to resolution.   
-- Right middle lobe pneumonia. 
\end{Verbatim}
    \end{minipage}
\end{figure}

\newpage
\subsection{Least Uncertain Instance at the Temperature
Setting of 0.25}

\begin{figure}[h]
    \centering
    \begin{minipage}{1.0\linewidth}
        \centering
        \includegraphics[width=8cm,height=9cm]
        {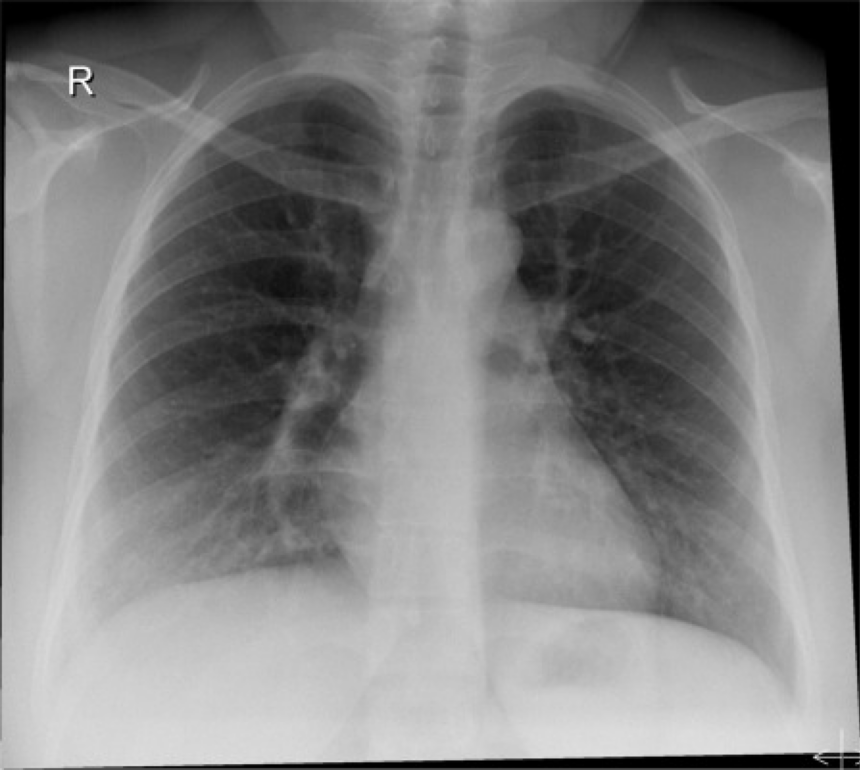}
        \caption*{X-Ray Image}
    \end{minipage}%
    \hfill
    \begin{minipage}{1.0\linewidth}
        \centering
\begin{Verbatim}
Image: 1-s2.0-S0378603X1500248X-gr3a.jpg

Responses:
-- 1-s2.0-S0378603X1500248X-gr3a.jpg
-- No acute cardiopulmonary process. 
-- 1.  Endotracheal tube terminates approximately 3.5 cm above the carina. 2.  Nasogastric tube courses below the diaphragm, terminating in the expected location of the stomach. 3.  Left upper lobe consolidation, which may represent aspiration or pneumonia.   
-- No acute cardiopulmonary process. 
-- No acute cardiopulmonary process. 
-- No acute cardiopulmonary process. 
-- No acute cardiopulmonary process. 
-- No acute cardiopulmonary process. 
-- No acute cardiopulmonary process. 
-- No acute cardiopulmonary process. 
-- No acute cardiopulmonary process. 
-- No acute cardiopulmonary process. 
-- No acute cardiopulmonary process. 
-- No acute cardiopulmonary process. 
-- No acute cardiopulmonary process. 
-- No acute cardiopulmonary process. 
-- No acute cardiopulmonary process. 
-- No acute cardiopulmonary process. 
-- No acute cardiopulmonary process. 
-- 1.  Endotracheal tube terminates 4.5 cm above the carina. 2.  Enteric tube courses below the diaphragm, out of the field of view. 3.  Diffuse bilateral airspace opacities may reflect pulmonary edema, aspiration, or infection.   
-- No acute cardiopulmonary process. 
-- No acute cardiopulmonary process. 
-- No acute cardiopulmonary process. 
-- 1.  Endotracheal tube terminates approximately 1.5 cm above the carina.  Recommend withdrawal by approximately 2 cm. 2.  Nasogastric tube courses below the diaphragm, terminating in the expected location of the stomach. 3.  Diffuse bilateral airspace opacities, which may reflect diffuse pulmonary edema or ARDS.  No pneumothorax.   
-- No acute cardiopulmonary process. 
-- No acute cardiopulmonary process. 
-- No acute cardiopulmonary process. 
-- No acute cardiopulmonary process. 
-- No acute cardiopulmonary process. 
\end{Verbatim}
    \end{minipage}
\end{figure}

\newpage
\subsection{Most Uncertain Instance at the Temperature Setting of 0.25}

\begin{figure}[h]
    \centering
    \begin{minipage}{1.0\linewidth}
        \centering
        \includegraphics[width=8cm,height=9cm]{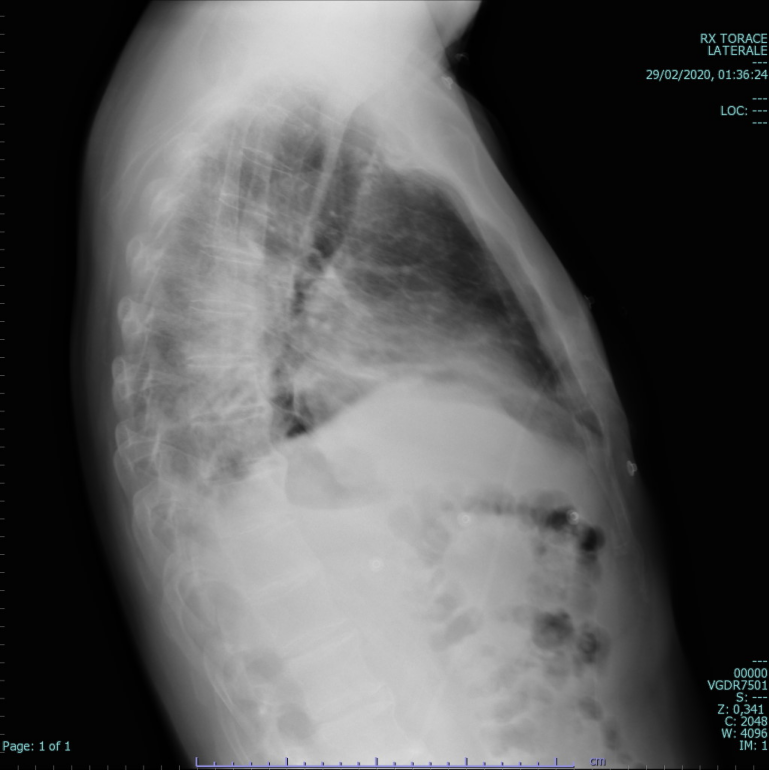}
        \caption*{X-Ray Image}
    \end{minipage}%
    \hfill
    \begin{minipage}{1.0\linewidth}
        \centering
\begin{Verbatim}
Image: 4C4DEFD8-F55D-4588-AAD6-C59017F55966.jpeg

Responses:
-- 4C4DEFD8-F55D-4588-AAD6-C59017F55966.jpeg
-- 1.  Interval placement of a right pigtail catheter with interval decrease in size of the right pleural effusion. 2.  No pneumothorax. 3.  Interval placement of a right internal jugular central venous catheter with tip in the mid SVC.   
-- 1.  Interval placement of a right-sided chest tube with associated decrease in size of right-sided pneumothorax. 2.  Right-sided volume loss and infiltrate.   
-- 1.  Interval placement of a right pigtail catheter with interval decrease in size of the right pleural effusion.  No pneumothorax. 2.  Worsening pulmonary edema. 3.  Right upper lobe collapse.   
-- 1.  Right upper lobe collapse. 2.  Moderate to large right pleural effusion. 3.  Left lower lobe collapse. 4.  Moderate to large left pleural effusion.   
-- 1.  Moderate right pleural effusion. 2.  Moderate pulmonary edema. 3.  Moderate cardiomegaly.   
-- 1.  Bilateral pleural effusions, moderate on the right and small on the left, with associated atelectasis. 2.  Right upper lobe collapse.   
-- 1.  Right upper lobe collapse. 2.  Moderate right pleural effusion. 3.  Moderate cardiomegaly.   
-- 1.  Right upper lobe collapse. 2.  Bilateral pleural effusions, left greater than right, with associated atelectasis. 3.  Moderate cardiomegaly.   
-- 1.  Bilateral pleural effusions, right greater than left, with associated bibasilar atelectasis. 2.  Moderate cardiomegaly.   
-- 1.  Bilateral pleural effusions, moderate on the right and small on the left. 2.  Right upper lobe collapse. 3.  Right lower lobe consolidation, likely atelectasis.   
-- 1.  Right upper lobe collapse. 2.  Bilateral pleural effusions. 3.  Moderately severe pulmonary edema.   
-- 1.  Interval placement of a right pigtail catheter with significant decrease in size of right pleural effusion. 2.  No evidence of pneumothorax. 3.  No evidence of pneumonia.   
-- 1.  Moderate right pleural effusion. 2.  Moderate cardiomegaly. 3.  No evidence of pneumonia.   
-- 1.  Bilateral pleural effusions, right greater than left, with associated atelectasis. 2.  Right upper lobe mass.  Recommend CT for further evaluation.   
-- 1.  No evidence of pneumothorax. 2.  Moderate-sized right pleural effusion. 3.  Extensive subcutaneous emphysema overlying the right chest wall.   
-- 1.  Interval placement of a right pigtail catheter with significant decrease in size of right pleural effusion. 2.  No pneumothorax. 3.  No evidence of pneumonia.   
-- 1.  Interval placement of a right pigtail catheter with interval decrease in size of the right pleural effusion. 2.  No pneumothorax. 3.  Interval placement of a right internal jugular central venous catheter with the tip in the mid SVC.   
-- 1.  Interval placement of a right pigtail catheter with interval decrease in size of the right pleural effusion. 2.  No pneumothorax. 3.  Worsening pulmonary edema.   
-- 1.  No evidence of pneumothorax. 2.  Bilateral pleural effusions, right greater than left, with associated atelectasis. 3.  Moderate cardiomegaly.   
-- 1.  Moderately-severe pulmonary edema. 2.  Bilateral pleural effusions, moderate to large on the right and small on the left. 3.  Right upper lobe collapse.   
-- 1.  Moderate right pneumothorax. 2.  Moderate right pleural effusion. 3.  Right upper lobe collapse.   
-- 1.  Right-sided chest tube in place with no evidence of pneumothorax. 2.  Bilateral parenchymal opacities, which may represent pulmonary contusion, aspiration, or pneumonia. 3.  Moderate-to-severe cardiomegaly.   
-- 1.  Right upper lobe collapse. 2.  Moderate right pleural effusion. 3.  Moderate cardiomegaly.   
-- 1.  Interval placement of a right pigtail catheter with interval decrease in size of the right pleural effusion. 2.  No pneumothorax. 3.  Worsening pulmonary edema.   
-- 1.  Bilateral pleural effusions, right greater than left, with associated bibasilar atelectasis. 2.  Diffuse osseous sclerosis, consistent with prostate cancer metastatic disease.   
-- 1.  Interval placement of a right-sided chest tube with persistent moderate-sized right apical pneumothorax. 2.  Right-sided rib fractures better assessed on the subsequent CT. 3.  Left basilar opacity may represent atelectasis or pneumonia.   
-- 1.  Bilateral pleural effusions, moderate in size on the right and small on the left, with associated bibasilar atelectasis. 2.  Moderate cardiomegaly.   
-- 1.  Interval placement of a right pigtail catheter with significant decrease in size of right pleural effusion. 2.  No pneumothorax. 3.  No evidence of pneumonia.   
\end{Verbatim}
    \end{minipage}
\end{figure}

\newpage
\subsection{Least Uncertain Instance at the Temperature
Setting of 0.50}
\begin{figure}[h]
    \centering
    \begin{minipage}{1.0\linewidth}
        \centering
        \includegraphics[width=8cm,height=9cm]      
        {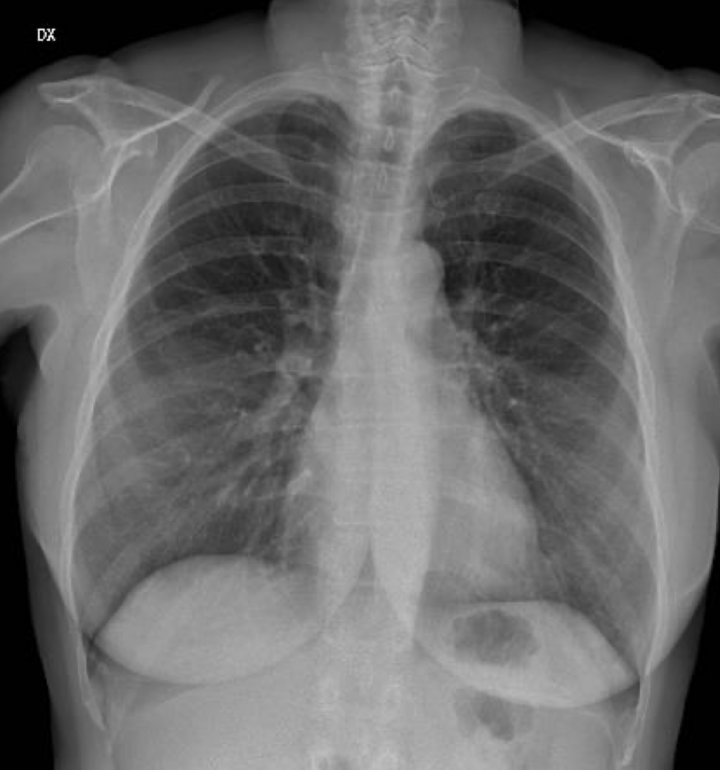}
        \caption*{X-Ray Image}
    \end{minipage}%
    \hfill
    \begin{minipage}{1.0\linewidth}
        \centering
\begin{Verbatim}
Image: 242a639e.jpg

Responses:
-- 242a639e.jpg
-- Findings concerning for multifocal pneumonia. 
-- No evidence of acute cardiopulmonary process. 
-- Findings concerning for pneumonia in the right lower lobe. 
-- Findings concerning for multifocal pneumonia. 
-- Findings suggestive of pneumonia in the right lower lobe. 
-- No acute intrathoracic process. 
-- 1.  Findings concerning for multifocal pneumonia. 2.  Recommend followup to resolution.   
-- Multifocal pneumonia.  Followup radiographs after treatment are recommended to ensure resolution of these findings.   
-- No acute intrathoracic process. 
-- Findings concerning for multifocal pneumonia. 
-- Right lower lobe pneumonia. 
-- Right lower lobe pneumonia. 
-- Findings concerning for multifocal pneumonia.  Recommend followup to resolution.   
-- Findings suggesting pneumonia in the lingula. 
-- Findings consistent with pneumonia in the lingula. 
-- No acute cardiopulmonary process. 
-- Findings compatible with pneumonia in the right lower lobe.  Recommend repeat after treatment to document resolution.   
-- Findings concerning for multifocal pneumonia. 
-- No acute cardiopulmonary process. 
-- Findings concerning for multifocal pneumonia.  Recommend followup to resolution.   
-- Findings suggestive of chronic lung disease with a component of infection not excluded.   
-- Findings compatible with pneumonia in the lingula. 
-- Findings suggesting chronic lung disease including bronchiectasis and airway wall thickening in the lingula.   
-- Findings suggestive of pneumonia in the right upper lobe. 
-- Findings compatible with pneumonia in the right lower lobe.  Recommend repeat after treatment to document resolution.   
-- Findings concerning for multifocal pneumonia. 
-- Findings suggesting pneumonia in the lingula. 
-- No acute cardiopulmonary process. 
\end{Verbatim}
    \end{minipage}
\end{figure}

\newpage
\subsection{Most Uncertain Instance at the Temperature
Setting of 0.50}

\begin{figure}[h]
    \centering
    \begin{minipage}{1.0\linewidth}
        \centering
        \includegraphics[width=8cm,height=9cm] {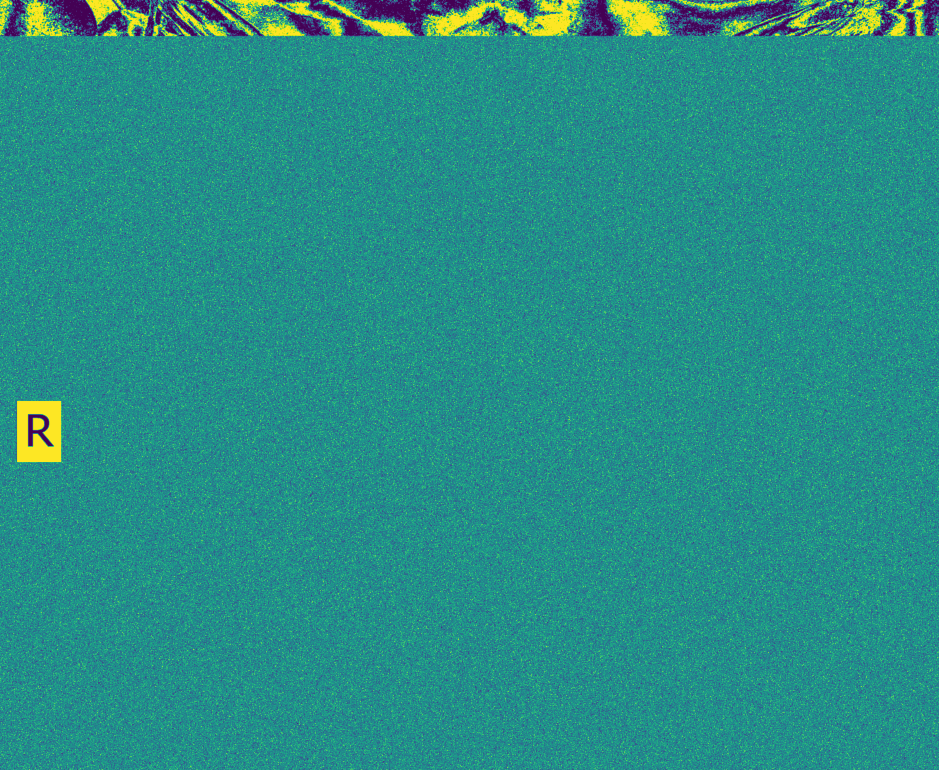}
        \caption*{X-Ray Image}
    \end{minipage}%
    \hfill
    \begin{minipage}{1.0\linewidth}
        \centering
\begin{Verbatim}
Image: 242a639e.jpg

Responses:
-- 242a639e.jpg
-- Findings concerning for multifocal pneumonia. 
-- No evidence of acute cardiopulmonary process. 
-- Findings concerning for pneumonia in the right lower lobe. 
-- Findings concerning for multifocal pneumonia. 
-- Findings suggestive of pneumonia in the right lower lobe. 
-- No acute intrathoracic process. 
-- 1.  Findings concerning for multifocal pneumonia. 2.  Recommend followup to resolution.   
-- Multifocal pneumonia.  Followup radiographs after treatment are recommended to ensure resolution of these findings.   
-- No acute intrathoracic process. 
-- Findings concerning for multifocal pneumonia. 
-- Right lower lobe pneumonia. 
-- Right lower lobe pneumonia. 
-- Findings concerning for multifocal pneumonia.  Recommend followup to resolution.   
-- Findings suggesting pneumonia in the lingula. 
-- Findings consistent with pneumonia in the lingula. 
-- No acute cardiopulmonary process. 
-- Findings compatible with pneumonia in the right lower lobe.  Recommend repeat after treatment to document resolution.   
-- Findings concerning for multifocal pneumonia. 
-- No acute cardiopulmonary process. 
-- Findings concerning for multifocal pneumonia.  Recommend followup to resolution.   
-- Findings suggestive of chronic lung disease with a component of infection not excluded.   
-- Findings compatible with pneumonia in the lingula. 
-- Findings suggesting chronic lung disease including bronchiectasis and airway wall thickening in the lingula.   
-- Findings suggestive of pneumonia in the right upper lobe. 
-- Findings compatible with pneumonia in the right lower lobe.  Recommend repeat after treatment to document resolution.   
-- Findings concerning for multifocal pneumonia. 
-- Findings suggesting pneumonia in the lingula. 
-- No acute cardiopulmonary process. 
\end{Verbatim}
    \end{minipage}
\end{figure}

\newpage
\subsection{Least Uncertain Instance at the Temperature
Setting of 0.75}

\begin{figure}[h]
    \centering
    \begin{minipage}{1.0\linewidth}
        \centering
        \includegraphics[width=8cm,height=9cm]{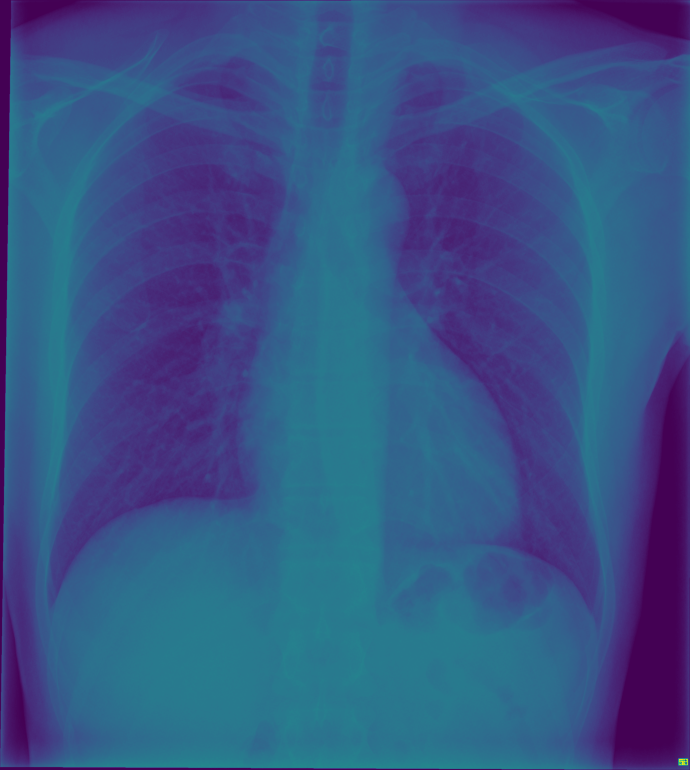}
        \caption*{X-Ray Image}
    \end{minipage}%
    \hfill
    \begin{minipage}{1.0\linewidth}
        \centering
\begin{Verbatim}
Image: 18017511.jpg

Responses:
-- 18017511.jpg
-- 1.  No radiographic evidence of pneumonia. 2.  Minimal right upper lobe linear atelectasis.   
-- No acute cardiopulmonary process. 
-- No radiographic evidence for acute cardiopulmonary process. 
-- No radiographic evidence of pneumonia. 
-- No acute cardiopulmonary process. 
-- No evidence of pneumonia. 
-- Normal chest radiograph. 
-- No evidence of acute cardiopulmonary process. 
-- No acute cardiopulmonary process. 
-- No acute cardiopulmonary process. 
-- No radiographic evidence of pneumonia. 
-- No acute cardiopulmonary process. 
-- No evidence of acute cardiopulmonary process. 
-- No acute cardiopulmonary process. 
-- No radiographic evidence for acute cardiopulmonary process. 
-- No acute cardiopulmonary process. 
-- No acute intrathoracic process. 
-- No radiographic evidence of pneumonia. 
-- Normal chest radiograph. 
-- Normal chest radiographs. 
-- No evidence of pneumonia. 
-- No acute intrathoracic process. 
-- No evidence of pneumonia. 
-- No acute cardiothoracic process. 
-- No acute cardiothoracic process. 
-- No acute cardiopulmonary process. 
-- No acute cardiothoracic process. 
-- No radiographic evidence of pneumonia. 
\end{Verbatim}
    \end{minipage}
\end{figure}

\newpage
\subsection{Most Uncertain Instance at the Temperature
Setting of 0.75}

\begin{figure}[h]
    \centering
    \begin{minipage}{1.0\linewidth}
        \centering
        \includegraphics[width=8cm,height=9cm]{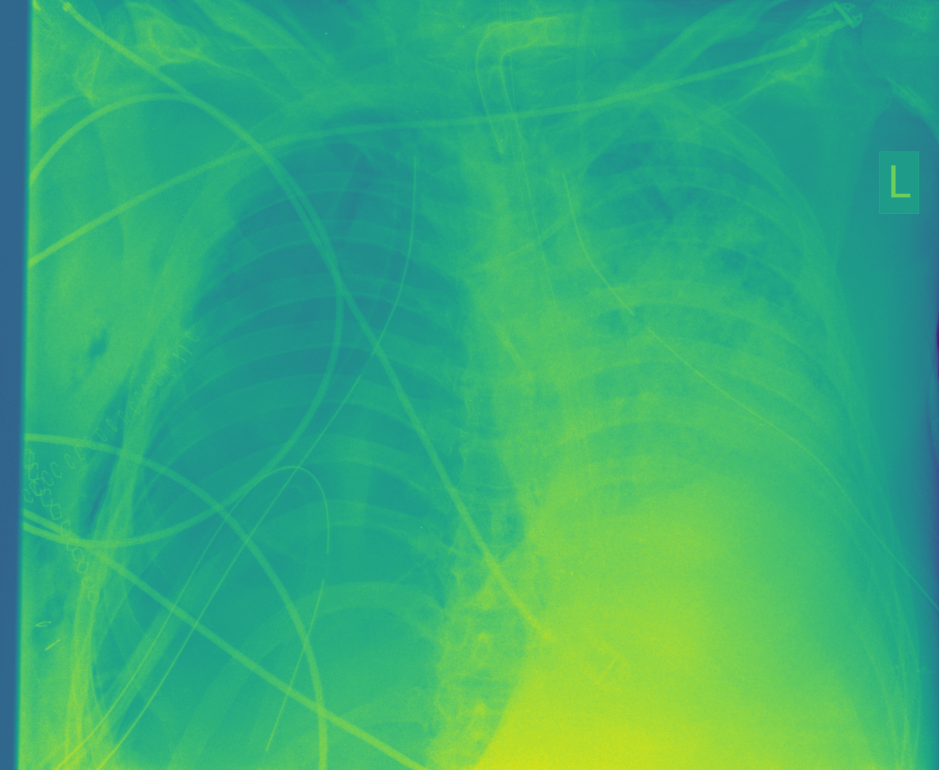}
        \caption*{X-Ray Image}
    \end{minipage}%
    \hfill
    \begin{minipage}{1.0\linewidth}
        \centering
\begin{Verbatim}
Image: 24de8686.jpg

Responses:
-- 24de8686.jpg
-- No acute cardiopulmonary process.  ET tube in appropriate position.   
-- No evidence of acute cardiopulmonary process. 
-- No acute cardiopulmonary process. 
-- No acute intrathoracic process. 
-- No acute cardiopulmonary abnormality. 
-- No acute cardiopulmonary process. 
-- No previous images.  The cardiac silhouette is within normal limits and there is no evidence of vascular congestion, pleural effusion, or acute focal pneumonia.   
-- 1.  No acute cardiothoracic process. 2.  No evidence of pneumomediastinum.   
-- AP chest compared to ___: Pulmonary vascular congestion and mild interstitial edema are new.  Atelectasis in the right lower lobe is mild.  Heart size is normal.  No pneumothorax or appreciable pleural effusion.   
-- No pneumothorax or other acute cardiopulmonary process. 
-- Heart size and mediastinum are stable.  Left basal consolidation has slightly increased.  There is minimal right basal atelectasis.  Right PICC line tip is at the level of lower SVC.  No pneumothorax is seen.   
-- No acute cardiopulmonary process. 
-- No evidence of acute cardiopulmonary process. 
-- No acute cardiopulmonary process. 
-- No acute cardiopulmonary process. 
-- No pneumothorax post biopsy. 
-- No acute cardiopulmonary process. 
-- No evidence of acute cardiopulmonary process. 
-- No acute cardiopulmonary process. 
-- 1. Right-sided Port-A-Cath terminates in the mid SVC. 2. Bibasilar atelectasis.   
-- No acute intrathoracic process. 
-- No acute cardiopulmonary process. 
-- 1.  Interval removal of endotracheal tube.  Enteric tube terminates below the diaphragm. 2.  Interval improvement of pulmonary edema. 3.  Right internal jugular line in the distal SVC.   
-- 1.  The nasogastric tube is coiled within the hiatal hernia, though the distal tip is within the stomach. 2.  The left subclavian central venous catheter tip terminates in the mid SVC. 3.  The endotracheal tube tip lies approximately 1.7 cm above the carina.  Recommend withdrawal by approximately 2.5 cm for more optimal positioning. 4.  There is a small left layering effusion.   
-- No acute intrathoracic process. 
-- In comparison with the study of ___, the monitoring and support devices are essentially unchanged.  Cardiac silhouette remains within normal limits.  The pulmonary vascularity is essentially within normal limits.  The hemidiaphragms are sharply seen with minimal if any pleural effusion.  No evidence of acute focal pneumonia.   
-- No evidence of pneumothorax. 
-- No acute cardiopulmonary process. 
\end{Verbatim}
    \end{minipage}
\end{figure}

\newpage
\subsection{Least Uncertain Instance at the Temperature
Setting of 1.00}

\begin{figure}[h]
    \centering
    \begin{minipage}{1.0\linewidth}
        \centering
        \includegraphics[width=8cm,height=9cm]{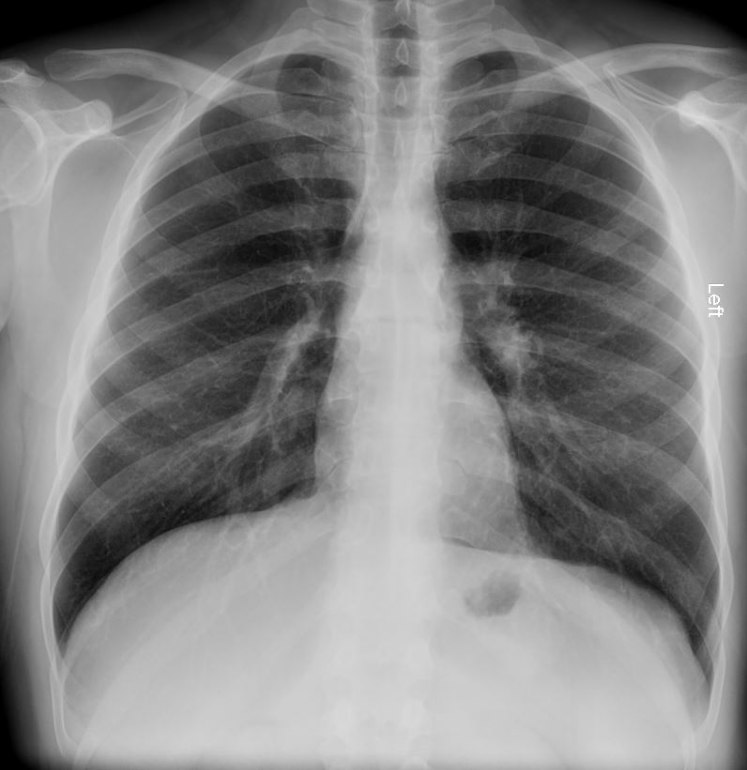}
        \caption*{X-Ray Image}
    \end{minipage}%
    \hfill
    \begin{minipage}{1.0\linewidth}
        \centering
\begin{Verbatim}
Image: covid-19-pneumonia-53.jpg

Responses:
-- covid-19-pneumonia-53.jpg
-- Lungs are fully expanded and clear.  Cardiomediastinal and hilar silhouettes and pleural surfaces are normal. .   
-- There no prior chest radiographs available for review. Heart size top-normal.  Pulmonary vasculature minimally more dilated but no focal pulmonary abnormality.  Pleural effusion small on the right if any.   
-- No acute cardiopulmonary process. 
-- 1.  Small right apical pneumothorax. 2.  Patchy and linear foci of atelectasis in the right lower lobe. 3.  NG tube ends in the stomach but the side port is near the gastroesophageal junction. This should be retracted by at least 7 cm for more optimal placement.   
-- Findings concerning for volume overload. 
-- Heart size is normal. Mediastinum is stable. Lungs are essentially clear. There is no pleural effusion or pneumothorax. Port-A-Cath catheter tip is at the level of mid SVC.   
-- New right pneumothorax with thickening signs of tension, right basal lateral pleural fluid and opacity likely post-inflammatory atelectasis.  Stable right upper lung opacity.   
-- Low lung volumes.  Moderate right-sided pneumothorax status post right-sided chest tube placement with significant decrease in size of pleural effusion.   
-- 1. No evidence of pneumonia or other acute cardiopulmonary process. 2. Stable rightward rotation of the heart.   
-- Unremarkable chest radiographic examination. 
-- No acute cardiopulmonary process. 
-- Bilateral linear opacities which are unchanged since ___ and may represent atypical infection.  Recommend followup to resolution.   
-- In comparison with the study of ___, there is little change in the appearance of the heart and lungs.  Specifically, no evidence of acute aspiration pneumonia.   
-- Mild pulmonary edema. 
-- Likely pneumonia within the lingula. 
-- No acute cardiopulmonary process. 
-- No evidence of free intraperitoneal air. 
-- No evidence of pulmonary edema. 
-- Normal chest radiographs 
-- Right PICC line tip in the region of the cavoatrial junction.  Please advance so that it is well within the SVC. NOTIFICATION:  Findings were communicated to Dr. ___ at 10:50 on ___ by phone.   
-- 1.  Right subclavian central venous catheter with tip in the mid-to-low SVC.  No pneumothorax. 2.  No acute cardiothoracic process.   
-- No acute cardiopulmonary process. 
-- 1. Increased interstitial opacities at the bases bilaterally and in the right upper lobe could represent pneumonia in the appropriate clinical setting. 2. Mild cardiomegaly with central pulmonary vascular prominence may be due to cardiac decompensation, and further evaluation with chest CT may be performed for confirmation and further characterization.   
-- New widespread parenchymal opacities, concerning for pneumonia. These findings were communicated via telephone by Dr. ___ to Dr. ___ at ___ on ___.   
-- Heart size and mediastinum are stable.  Right PICC line tip is at the level of mid SVC.  Lungs are overall clear.  No pleural effusion or pneumothorax is seen.   
-- Bibasilar opacities may be due to atelectasis versus aspiration. 
-- Right lower lobe pneumonia. 
-- Minor atelectasis of right middle lobe.  Otherwise, no acute cardiopulmonary process.   
\end{Verbatim}
    \end{minipage}
\end{figure}

\newpage
\subsection{Most Uncertain Instance at the Temperature
Setting of 1.00}

\begin{figure}[h]
    \centering
    \begin{minipage}{1.0\linewidth}
        \centering
        \includegraphics[width=8cm,height=9cm]{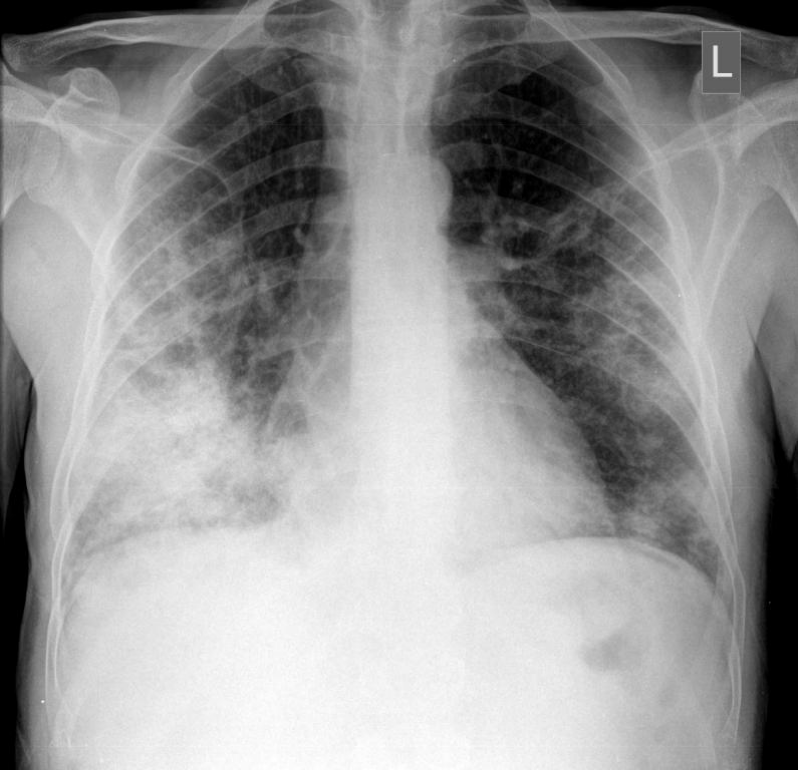}
        \caption*{X-Ray Image}
    \end{minipage}%
    \hfill
    \begin{minipage}{1.0\linewidth}
        \centering
\begin{Verbatim}
Image: 000004.jpg

Responses:
-- 000004.jpg
-- AP chest reviewed in the absence of prior chest radiographs: Lung volumes are very low, exaggerating cardiomediastinal caliber and crowding of pulmonary vasculature.  There is mild central pulmonary vascular engorgement, but no pulmonary edema, focal pulmonary consolidation, or appreciable pleural effusion. ET tube in standard placement.  Nasogastric tube passes into the stomach and out of view.  No pneumothorax.  Healed rib fractures noted.   
-- 1.  Enteric tube is seen coursing below the diaphragm, out of the field of view. 2.  ETT is seen in the appropriate position 3.  Limited evaluation due to underlying trauma and rotation.  Please refer to subsequent CT of the neck and chest for further details 4.  Extensive left upper lobe consolidative opacity which may represent pneumonia or pulmonary hemorrhage depending on the clinical scenario 5.  Small pleural effusion on the left   
-- Nasogastric tube tip in the stomach.  Right internal jugular central venous line at the cavoatrial junction.   
-- Pulmonary edema with moderate right-sided effusion. 
-- ET tube 5.3 cm above the carina. 
-- This is an AP chest reviewed in the absence of prior chest radiographs: Nasogastric tube ends in the stomach.  ET tube in standard placement.  Midline and left pleural drains in place.  Heterogeneous opacification of the lungs and substantial volume loss in both upper lungs is most commonly due to large scale aspiration.  Heart is mildly enlarged.  There is no pneumothorax. ET tube in standard placement.  Nasogastric tube in the stomach.   
-- AP chest reviewed in the absence of prior chest radiographs: Mild interstitial edema has worsened, moderately severe pulmonary edema has increased, and a moderate right pleural effusion is presumed.  Heart size is top normal.  No pneumothorax.  ET tube tip at the upper margin of the clavicles, no less than 3 cm from the carina. Nasogastric tube passes below the diaphragm and out of view.  No pneumothorax.   
-- Complete opacification of both hemithoraces likely secondary to bilateral pneumonic infiltrates.  Infiltrate in the right upper lobe also possible. These findings were discussed via telephone by Dr. ___ with Dr. ___ at ___ on ___.   
-- 1.  Interval placement of an endotracheal tube in appropriate position. 2.  Interval development of multifocal airspace consolidation, predominantly in the right upper and lower lobes and left perihilar region which may represent asymmetrical pulmonary edema or multifocal infection.  Consider repeat radiographs after diuresis to help further assess. 3.  Heart appears enlarged.  Attention to this in the follow-up imaging would be helpful to exclude the possibility of underlying pericardial effusion.  Recommendations discussed with Dr. ___ at 11:50 on ___ via telephone.   
-- Since ___, asymmetrically distributed moderate pulmonary edema with bilateral perihilar and basilar opacities, has worsened, moderate cardiomegaly, and a small right pleural effusion.   
-- Concern for aspiration/pulmonary hemorrhage. 
-- Bilateral effusions, pulmonary vascular congestion, moderate pulmonary edema, and left lung collapse.   
-- Complete opacification of the right hemithorax with just a small amount of aeration at the medial right apex, likely due to underlying consolidation and/or effusion.  Worsening vascular plethora and diffuse alveolar opacities which could be due to multifocal aspiration and/or evolving pulmonary edema.   
-- Complete opacification of both hemithoraces 
-- Status post endotracheal intubation.  New diffuse hazy opacity in the right lung, which should be followed to resolution to exclude aspiration.   
-- Extensive consolidation in the right upper and lower lobe, likely multi focal pneumonia. Recommend followup to resolution.  Bilateral moderate pulmonary edema.   
-- 1.  Nondisplaced right 1st rib fracture. 2.  Diffuse pneumonia. 3.  Cardiomediastinal and hilar widening and enlargement, possibly due to cardiomegaly and central lymph node enlargement.  Correlation with radiographs performed at or near the time of radiography is recommended to confirm that this is stable rather than an acute change.   
-- Diffuse moderately severe pulmonary edema with probable background COPD.  Infection, pulmonary contusion or aspiration in the appropriate clinical situation cannot be excluded.   
-- Near-complete opacification of the right hemithorax with just a small amount of aeration seen.  This is suggestive of near-complete collapse of the right lung with probable underlying atelectasis or consolidation.   
-- Endotracheal tube terminates 4 cm above the carina. Orogastric tube is below the diaphragm, the tip not visualized.   
-- Endotracheal tube projects 2.7 cm above the carinal.  Enteric tube side port located in the distal esophagus and must be advanced. Left perihilar opacity which may be secondary to aspiration versus atelectasis versus pneumonia. NG tube side port in the distal esophagus and must be advanced.   
-- In comparison to ___ chest radiograph, marked enlargement of the cardiac silhouette is accompanied by improved pulmonary edema and near resolution of small pleural effusions. .  Widespread intrathoracic malignancy is poorly imaged on this portable radiograph.  Extensive subcutaneous emphysema is present throughout the imaged neck and right chest wall. NOTIFICATION:  The findings were discussed by Dr. ___ with Dr. ___ on the telephone on ___ at 9:20 AM, 15 minutes after discovery of the findings.   
-- 1.  Complete opacification of the right hemithorax, concerning for a large right pleural effusion, possibly malignant. 2.  Significant vascular enlargement.   
-- The ET tube terminates 4 cm above the carina.  Right IJ line terminates in the low SVC.  Left-sided pacemaker and leads are in appropriate position.   
-- Endotracheal tube terminates 4.8 cm above the carina.  Nasogastric tube terminates at the gastroesophageal junction and should be advanced. Perihilar and diffuse alveolar opacities concerning for pulmonary edema, superimposed infection, and cardiac decompensation.   
-- Endotracheal tube positioned appropriately.  Extremely extensive left lung opacification concerning for hemorrhage.  Extensive pneumonia, possibly multifocal aspiration.  Enteric tube positioned appropriately.   
-- 1.  ETT and NG tubes in appropriate position. 2.  Severe pulmonary edema. 3.  Complete opacification of both hemithoraces, likely representing bilateral layering pleural effusions and associated atelectasis.   
-- 1.  Status post endotracheal intubation. 2.  Bilateral perihilar opacities concerning for pulmonary edema.   
\end{Verbatim}
    \end{minipage}
\end{figure}

\end{document}